\pdfoutput=1
\documentclass{article} 
\usepackage{iclr2027_conference,times}

\usepackage{amsmath,amsfonts,bm}

\def\eqref#1{equation~\ref{#1}}
\def\Eqref#1{Equation~\ref{#1}}

\def\1{\bm{1}}

\DeclareMathAlphabet{\mathsfit}{\encodingdefault}{\sfdefault}{m}{sl}
\SetMathAlphabet{\mathsfit}{bold}{\encodingdefault}{\sfdefault}{bx}{n}

\usepackage{url}
\usepackage{amsmath,amssymb}
\usepackage{dsfont} 
\usepackage{booktabs}
\usepackage{multirow}
\usepackage{graphicx}
\usepackage{xcolor}
\usepackage{enumitem}
\usepackage{rotating} 
\usepackage{tikz}
\usetikzlibrary{arrows.meta,positioning,fit}
\usepackage{pgfplots}
\pgfplotsset{compat=1.17}
\usepackage{placeins} 
\usepackage[colorlinks=true,linkcolor=blue!60!black,citecolor=blue!60!black,urlcolor=blue!60!black]{hyperref} 
\makeatletter
\g@addto@macro\normalsize{\setlength{\abovedisplayskip}{5pt plus 1pt minus 2pt}\setlength{\belowdisplayskip}{5pt plus 1pt minus 2pt}\setlength{\abovedisplayshortskip}{2pt}\setlength{\belowdisplayshortskip}{3pt}}
\makeatother
\renewcommand{\paragraph}[1]{\par\vspace{2.5pt plus 1pt minus 1pt}\noindent\textbf{#1}\hspace{0.55em}\ignorespaces}
\newcommand{\spp}{SelPI}                        
\newcommand{\sppone}{\spp{}$_{\mathrm{1G}}$}    
\newcommand{\spptwo}{\spp{}$_{\mathrm{2G}}$}    
\newcommand{\gpi}{g_{\mathrm{PI}}}
\newcommand{\gval}{g_{\mathrm{val}}}
\newcommand{\ind}{\mathds{1}}

\newcommand{\picp}{\mathrm{PICP}}
\newcommand{\mpiw}{\mathrm{MPIW}}
\newcommand{\IS}{\mathrm{IS}}
\newcommand{\aurc}{\mathrm{AURC}}

\newcommand{\yhat}{\hat{y}}
\newcommand{\phit}{\phi_{\mathrm{t}}}

\title{Selective Prediction Intervals: Jointly Learning to Predict, Bound, and Abstain}

\author{Lidor Rubi \\
Faculty of Computer and Information Science\\
Ben-Gurion University of the Negev\\
Beer-Sheva, Israel \\
\texttt{rubil@post.bgu.ac.il}
\And
Gilad Katz \\
Faculty of Computer and Information Science\\
Ben-Gurion University of the Negev\\
Beer-Sheva, Israel \\
\texttt{giladkz@bgu.ac.il}
}

\iclrfinalcopy
\begin{document}

\maketitle
\lhead{Preprint. Under review.}

\begin{abstract}
Many high-stakes regression decisions ask three things of a model at once: a point estimate to act on, an interval that bounds the risk of acting, and the option to abstain when neither can be trusted. Deep learning provides these in isolation: prediction-interval methods return an interval, and at best a value estimate inside it, for \emph{every} input, while selective-prediction methods learn when to abstain, but only for point predictions and with no statement of how far the truth may lie from the estimate. We introduce \spp{}, which trains interval, value and abstention heads jointly under a single loss with an explicit acceptance-rate budget. A natural objection is that interval width already tells us when to abstain. We show that it does not: across 18 tabular benchmarks, width predicts large point errors well (AUROC $0.72$) but is \emph{anti}-correlated with interval misses (AUROC $0.38$), so abstaining on wide intervals discards precisely the inputs the interval got right. \spp{} therefore selects with two gates, one trained on the interval loss and one on the value loss, coupled by a Fr\'echet penalty that ties their separate budgets to the joint acceptance rate they deliver. With the base predictor held fixed, the learned gate beats width thresholding on a proper interval score on all 18 tabular datasets, closing a median $87\%$ of the gap to an oracle selector, and it also beats selection by deep-ensemble disagreement on every dataset. The advantage grows with the heteroscedasticity of the task, and we report the settings in which it disappears.
\end{abstract}

\section{Introduction}
\label{sec:intro}

Consider a pediatric radiologist using a model to estimate skeletal age from a hand radiograph. A bare point estimate is not enough: the clinical question is whether bone age is \emph{consistent} with chronological age, which requires a plausible range. A range alone is not enough either, because the report must state a value. And for some radiographs---unusual poses, occlusions, rare conditions---the honest output is a referral for manual reading. The same triad appears wherever a regression model feeds a decision: these are components of a single decision, and a model that provides only some of them pushes the missing one onto the user.

Two lines of research supply these components separately. \emph{Prediction-interval} (PI) methods learn a lower and an upper bound around the target~\citep{khosravi2010lower,pearce2018high}; IPIV~\citep{simhayev2022integrated} additionally learns \emph{where inside the interval} the target is likely to be, yielding a point estimate coherent with its own bounds. These methods answer every query. \emph{Selective-prediction} methods, exemplified by SelectiveNet~\citep{geifman2019selectivenet}, learn a gate that abstains on a budgeted fraction of inputs so that the risk on the accepted remainder falls; they return only a point estimate. Neither family delivers the full triad: the accepted set of a selective point predictor carries no statement of risk, and the intervals of a non-selective predictor are only as trustworthy as the model's worst inputs.

The obvious way to close the gap is post hoc: train an interval model and abstain when the interval is wide. Understanding why this baseline falls short is the core of this paper. Width is an \emph{aleatoric} signal, largely blind to \emph{epistemic} failure---sparsely covered regions, distribution shift, intervals that are confidently \emph{wrong}. And when the point estimate lies inside the interval by construction, point error is mechanically bounded by width on every input the interval captures (\S\ref{sec:background}). The failures that matter are therefore the two that width cannot bound: the interval \emph{misses} the target, or it captures the target but places the point estimate at the wrong end of a wide interval.

We propose \spp{} (\emph{Sel}ective \emph{P}rediction \emph{I}ntervals), which trains IPIV's interval and value heads jointly with SelectiveNet-style gates under a single selective interval loss with an explicit acceptance budget. Selection is split across two gates---$\gpi$, trained on the selective interval loss and therefore driven by miscoverage, and $\gval$, trained on the selective value loss and therefore driven by value-position error---which must \emph{both} admit an input for it to be answered. Each gate has its own budget, and a Fr\'echet penalty ties the two budgets to the joint acceptance rate they actually deliver.

Because no existing method produces the same output, the evaluation straddles two fields, and we organize it as three questions (\S\ref{sec:setup}, Table~\ref{tab:design}): does learning to abstain damage the interval model (the PI field's question); is the result a competitive selective regressor (the selective-prediction field's question); and, holding the predictor fixed, does a learned gate select better than the predictor's own uncertainty (the new question). The answer to the third is a clear yes; the answers to the first two are ``mostly'', and we report the exceptions. Our contributions are:
\begin{enumerate}[leftmargin=*,itemsep=0pt,topsep=1pt]
  \item \spp{}, the first method to jointly learn prediction intervals, an in-interval value estimate, and a budgeted abstention rule. We identify the structural coupling between an in-interval value estimate and interval width, and derive from it a \emph{dual-gate} selector whose heads specialize to the two failure modes width cannot detect (\S\ref{sec:method}).
  \item An evaluation protocol for selective interval prediction: selective risk curves for a proper interval score, a same-predictor comparison that isolates the selector from the base model, and predictor-independent gate diagnostics (\S\ref{sec:setup}).
  \item A characterization of \emph{when} learned selection helps: the gain over width thresholding grows with residual heteroscedasticity and vanishes on near-deterministic problems, and each gate's rejections are enriched for the failure mode it was trained on (\S\ref{sec:analysis}).
\end{enumerate}

\section{Related Work}
\label{sec:related}

\paragraph{Selective prediction.}
Learning to abstain dates to Chow's error--reject trade-off~\citep{chow1970optimum}. \citet{el2010foundations} introduced the risk--coverage curve, and \citet{cortes2016learning} gave the problem a learning-theoretic treatment. Thresholding the softmax response is a strong post-hoc selector for deep classifiers~\citep{geifman2017selective}. SelectiveNet~\citep{geifman2019selectivenet} instead trains the selector jointly with the predictor under an explicit coverage budget, with an auxiliary head that keeps the representation from collapsing onto the accepted subset. Later work recasts abstention as betting~\citep{liu2019deep}, uses a Gumbel-softmax gate~\citep{salem2022gumbel}, adds entropy regularizers~\citep{feng2022towards}, or calibrates the accepted set~\citep{fisch2022calibrated}. The excess-AURC of \citet{geifman2018bias} isolates selector quality from base-model quality and underlies our main statistic. Selective \emph{regression} is comparatively unexplored. Apart from a few regression experiments in SelectiveNet and its Gumbel-softmax variant, existing methods select post hoc, by thresholding an uncertainty score of a trained model. That score is the conditional variance, which is the optimal rule for squared loss~\citep{zaoui2020regression}, a blended variance with a risk guarantee~\citep{jiang2020risk}, model-agnostic bootstrap uncertainty~\citep{pugnana2023model}, a conformal confidence score~\citep{sokol2026conformalized}, or the CRPS entropy of a distributional forecast~\citep{zaoui2025distributional}. \citet{cheng2023regression} study rejection under an explicit cost and \citet{shah2022selective} under fairness constraints. All of these select on a single scalar, and none learns an interval jointly with the selector; our width-threshold and ensemble-disagreement baselines are this family applied to an interval model.

\paragraph{Prediction intervals for deep regression.}
Distribution-free interval learning began with LUBE~\citep{khosravi2010lower}, whose coverage--width objective QD~\citep{pearce2018high} made differentiable. QD+~\citep{salem2020prediction} added loss terms for a point prediction and a split-normal mixture for aggregation. IPIV~\citep{simhayev2022integrated} obtains the point prediction from a value head instead, yielding a coherent (point, interval) pair without extra loss terms. Quantile regression~\citep{koenker1978regression,tagasovska2019single,chung2021beyond} and heteroscedastic likelihoods~\citep{nix1994estimating,kendall2017uncertainties} are alternative routes to intervals. Epistemic uncertainty is usually estimated with MC dropout~\citep{gal2016dropout} or deep ensembles~\citep{lakshminarayanan2017simple}, and \citet{hullermeier2021aleatoric} review the aleatoric/epistemic distinction that organizes our analysis. Conformal prediction~\citep{vovk2005algorithmic,romano2019conformalized,angelopoulos2023conformal} wraps any of these with a marginal finite-sample guarantee, and recent work calibrates prediction \emph{sets} on a post-hoc selected subset~\citep{xu2025selective,kumar2025beyond}. All of these either answer every query or take the selector as given. We learn the selector jointly with the intervals (extended discussion in Appendix~\ref{app:related}).

\section{Background}
\label{sec:background}

\paragraph{Setup and terminology.}
We observe $(x, y) \in \mathcal{X} \times \mathbb{R}$ drawn i.i.d.\ from an unknown $P$. A \emph{prediction interval} at nominal miscoverage $\alpha$ is a pair $L, U : \mathcal{X} \to \mathbb{R}$ intended to satisfy $\Pr[L(x) \le y \le U(x)] \ge 1 - \alpha$; its empirical quality is summarized by the coverage probability $\picp = \frac{1}{n}\sum_i k_i$, with $k_i = \ind[L(x_i) \le y_i \le U(x_i)]$ the \emph{capture} indicator, and the mean width $\mpiw = \frac{1}{n}\sum_i (U(x_i) - L(x_i))$. A \emph{selective predictor} is a pair $(f, g)$ with gate $g : \mathcal{X} \to [0,1]$ that answers $f(x)$ when $g(x) \ge \tau$ and abstains otherwise. For a loss $\ell$, its \emph{selective risk}~\citep{el2010foundations} is the average loss over the accepted subset,
\begin{equation}
  R_\ell(f, g; \tau) \;=\; \frac{\sum_i \ell(f(x_i), y_i)\,\ind[g(x_i) \ge \tau]}{\sum_i \ind[g(x_i) \ge \tau]} .
  \label{eq:selrisk}
\end{equation}
``Coverage'' means accepted inputs in the selective literature and captured targets in the PI literature, so we never use the word unqualified: the former is the \emph{acceptance rate} $\phi$, the latter $\picp$. $\tau_\phi$ is the threshold that admits a fraction $\phi$ of inputs, and $R_\ell(\phi)$ abbreviates $R_\ell(f,g;\tau_\phi)$.

\paragraph{SelectiveNet.}
SelectiveNet~\citep{geifman2019selectivenet} attaches a predictor $f$, a soft gate $g$ and an auxiliary predictor $h$ to a shared backbone, and minimizes a gate-weighted risk under a soft acceptance constraint,
\begin{equation}
  \mathcal{L}_{(f,g)} = \frac{\frac{1}{n}\sum_i \ell(f(x_i), y_i)\,g(x_i)}{\hat{\phi}(g)} + \lambda \,\Psi\big(\phit - \hat{\phi}(g)\big),
  \qquad \hat{\phi}(g) = \tfrac{1}{n}\textstyle\sum_i g(x_i),
  \label{eq:selectivenet}
\end{equation}
with $\phit$ the target acceptance rate and $\Psi(a) = \max(0,a)^2$. The auxiliary head is trained on \emph{all} inputs with the plain loss $\mathcal{L}_h$ and discarded at test time; the objective $\gamma\,\mathcal{L}_{(f,g)} + (1-\gamma)\,\mathcal{L}_h$ keeps the backbone from overfitting the accepted subset. One model is trained per $\phit$.

\paragraph{IPIV.}
IPIV~\citep{simhayev2022integrated} outputs bounds $U(x), L(x)$ and a mixing coefficient $v(x) \in (0,1)$, which we call the \emph{value head},\footnote{IPIV calls $v$ its ``auxiliary head''. We reserve \emph{auxiliary} for SelectiveNet's all-input heads to avoid a clash.} and predicts
\begin{equation}
  \yhat(x) \;=\; v(x)\,U(x) + \big(1 - v(x)\big)\,L(x).
  \label{eq:piven-point}
\end{equation}
The bounds are trained with the quality-driven objective of QD~\citep{pearce2018high}: minimize the width of the intervals that capture their target, $\mpiw_{\mathrm{capt}} = \sum_i k_i (U_i - L_i) / \sum_i k_i$, under a penalty that activates when coverage falls below nominal,
\begin{equation}
  \mathcal{L}_{\mathrm{PI}} = \mpiw_{\mathrm{capt}} + \sqrt{n}\,\lambda\,\Psi\big((1-\alpha) - \widetilde{\picp}\big),
  \label{eq:qd}
\end{equation}
where $\widetilde{\picp}$ replaces the hard capture indicator with the sigmoid relaxation $\tilde{k}_i = \sigma\big(s(U_i - y_i)\big)\,\sigma\big(s(y_i - L_i)\big)$, and $n$ is the batch size (Appendix~\ref{app:loss}). The value head is trained with a regression loss on \eqref{eq:piven-point}, $\mathcal{L}_{v}$ (squared error here), on \emph{all} inputs---captured or not---and the total loss is $\beta\,\mathcal{L}_{\mathrm{PI}} + (1-\beta)\,\mathcal{L}_{v}$. IPIV motivates this design in the same way SelectiveNet motivates its auxiliary head: a loss that only sees the captured subset overfits to it.

\paragraph{A structural coupling.}
\Eqref{eq:piven-point} implies $\yhat(x) \in [L(x), U(x)]$, so on any captured input $|\yhat_i - y_i| \le U_i - L_i$: \emph{point error is bounded by interval width wherever the interval is correct}. Two consequences shape the rest of the paper. First, abstaining on wide intervals reduces selective point error automatically, so a width threshold is an unusually strong baseline on point-error curves. Second, a learned gate can genuinely separate from width only on the events width does not bound: a \emph{coverage failure} ($y_i \notin [L_i, U_i]$) and a \emph{value failure} ($y_i$ captured, but $v_i$ far from the target's true relative position $v_i^\star = (y_i - L_i)/(U_i - L_i)$). Section~\ref{sec:method} builds the selector around exactly these two events.

\section{\spp{}: Selective Prediction Intervals}
\label{sec:method}

\begin{figure}[t]
\centering
\begin{tikzpicture}[
  font=\scriptsize,
  >={Stealth[length=1.5mm]},
  box/.style={draw, rounded corners=1.5pt, minimum height=4.6mm, inner sep=1.5pt, align=center},
  hd/.style={box, minimum width=6.4mm},
  lab/.style={font=\scriptsize\itshape, inner sep=0pt, anchor=south west, yshift=1.3pt},
  bar/.style={line width=0.9pt, {|[width=5pt]}-{|[width=5pt]}},
]
\node[box, fill=gray!10] (x) at (0,0) {input\\$x$};
\node[box, fill=gray!25, minimum width=13mm, minimum height=27mm] (bb) at (1.55,0) {shared\\backbone};
\draw[->] (x) -- (bb);
\node[hd, fill=blue!12] (L) at (3.35,0.95) {$L$};
\node[hd, fill=blue!12] (U) at (4.10,0.95) {$U$};
\node[hd, fill=blue!12] (v) at (4.85,0.95) {$v$};
\node[draw=blue!50, rounded corners=2pt, inner sep=2pt, fit=(L)(v)] (G1) {};
\node[lab] at (G1.north west) {interval and value heads};
\node[hd, fill=orange!22, minimum width=9mm] (gpi) at (3.50,-0.05) {$g_{\mathrm{PI}}$};
\node[hd, fill=orange!22, minimum width=9mm] (gval) at (4.70,-0.05) {$g_{\mathrm{val}}$};
\node[draw=orange!70, rounded corners=2pt, inner sep=2pt, fit=(gpi)(gval)] (G2) {};
\node[lab] at (G2.north west) {selection gates};
\node[hd, fill=gray!8, dashed] (Lh) at (3.35,-1.05) {$L_h$};
\node[hd, fill=gray!8, dashed] (Uh) at (4.10,-1.05) {$U_h$};
\node[hd, fill=gray!8, dashed] (vh) at (4.85,-1.05) {$v_h$};
\node[draw=gray, dashed, rounded corners=2pt, inner sep=2pt, fit=(Lh)(vh)] (G3) {};
\node[lab] at (G3.north west) {auxiliary heads};
\foreach \g in {G1,G2,G3} \draw[->] (bb.east |- \g.west) -- (\g.west);
\node[box, fill=blue!5, minimum width=19mm] (out) at (6.75,0.95) {interval $[L,U]$\\point $\hat{y}$ inside it};
\node[box, fill=orange!8, minimum width=19mm] (dec) at (6.75,-0.05) {both accept?\\answer / abstain};
\node[align=center, text=gray!70!black] (auxt) at (6.75,-1.05) {all inputs, training\\only; regularizer};
\draw[->] (G1.east) -- (out.west);
\draw[->] (G2.east) -- (dec.west);
\draw[->, gray, dashed] (G3.east) -- (auxt.west);
\draw[->] (dec.north) -- (out.south);
\draw[gray!50] (8.05,-1.45) -- (8.05,1.55);
\node[lab, anchor=south west] at (8.3,1.30) {what interval width cannot tell apart};
\draw[bar, blue!60!black] (8.35,0.95) -- (10.55,0.95);
\fill (9.35,0.95) circle (1.5pt);
\node[font=\scriptsize] at (9.60,0.95) {\textcolor{red!75!black}{$\bigstar$}};
\node[anchor=west, align=left, text width=27.5mm] at (10.75,0.95) {wide but right: a width rule rejects, both gates accept};
\draw[bar, blue!60!black] (8.95,-0.05) -- (9.75,-0.05);
\fill (9.35,-0.05) circle (1.5pt);
\node[font=\scriptsize] at (10.30,-0.05) {\textcolor{red!75!black}{$\bigstar$}};
\node[anchor=west, align=left, text width=27.5mm] at (10.75,-0.05) {narrow but misses $y$: only $g_{\mathrm{PI}}$ rejects};
\draw[bar, blue!60!black] (8.35,-1.05) -- (10.55,-1.05);
\fill (8.65,-1.05) circle (1.5pt);
\node[font=\scriptsize] at (10.30,-1.05) {\textcolor{red!75!black}{$\bigstar$}};
\node[anchor=west, align=left, text width=27.5mm] at (10.75,-1.05) {captured, $\hat{y}$ at the wrong end: $g_{\mathrm{val}}$ rejects};
\end{tikzpicture}
\caption{\textbf{\spp{} in one picture.} \emph{Left:} one backbone feeds the interval and value heads, two selection gates, and auxiliary heads that are trained on all inputs and dropped at inference; a prediction is released only if \emph{both} gates accept. \emph{Right:} the cases a width threshold confuses (bar: predicted interval; dot: point estimate $\hat{y}$; star: true $y$).}
\label{fig:arch}
\end{figure}
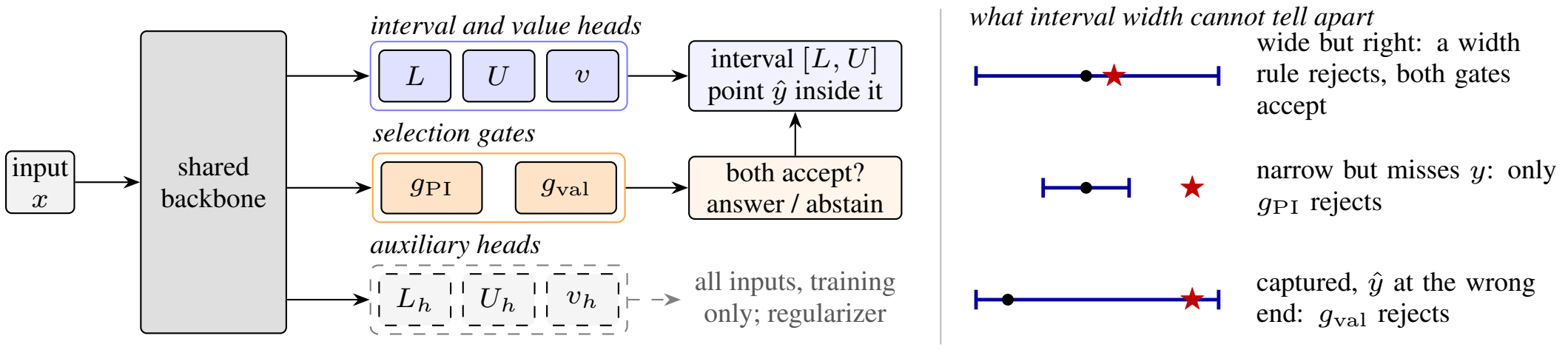

\subsection{Architecture}
\label{sec:arch}
\spp{} attaches three groups of outputs to a shared backbone (Figure~\ref{fig:arch}): the IPIV triple $\big(L(x), U(x), v(x)\big)$ with $\yhat(x)$ as in \eqref{eq:piven-point}; two gates $\gpi(x), \gval(x) \in [0,1]$; and an auxiliary triple $\big(L_h(x), U_h(x), v_h(x)\big)$ that is trained on all inputs and discarded at inference. The backbone and the $(L,U,v)$ heads are IPIV's for the task, the auxiliary triple duplicates the heads, and each gate is SelectiveNet's selection head (Appendix~\ref{app:hyper}). The auxiliary triple plays the role of SelectiveNet's auxiliary head (\S\ref{sec:background}): it is trained on every input, so the shared backbone keeps learning from the inputs the gates reject rather than overfitting the accepted subset. Its loss is given in \S\ref{sec:single} and its contribution is measured in \S\ref{sec:ablations}. \spp{} therefore adds only the gates and the auxiliary triple to an IPIV model, which is what makes the full-acceptance comparison in \S\ref{sec:results} a fair test of what the additions cost.

\subsection{Selective interval and value losses}
\label{sec:single}
We instantiate SelectiveNet's gate-weighted risk~\eqref{eq:selectivenet} not with a point loss but with IPIV's two objectives, so that both the interval and the value estimate are optimized on the inputs the model will actually answer. For a gate $g$ with $g_i = g(x_i)$, define the gate-weighted capture rate and captured width,
\begin{equation}
  \picp_g = \frac{\sum_i g_i\,\tilde{k}_i}{S_g}, \qquad
  \mpiw_g = \frac{\sum_i g_i\,k_i\,(U_i - L_i)}{\sum_i g_i\,k_i}, \qquad
  S_g = \textstyle\sum_i g_i, \qquad
  \hat{\phi}(g) = S_g / n ,
  \label{eq:gated-stats}
\end{equation}
where $S_g$ is the effective number of selected samples. The \emph{selective interval loss} carries both constraints---interval coverage and acceptance rate---and the \emph{selective value loss} is the gate-weighted squared error:
\begin{align}
  \mathcal{L}^{\mathrm{sel}}_{\mathrm{PI}}(g) &= \mpiw_g
    + \sqrt{S_g}\,\lambda_{\mathrm{PI}}\,\Psi\big((1-\alpha) - \picp_g\big)
    + \lambda_{\mathrm{sel}}\,\Psi\big(\phit - \hat{\phi}(g)\big),
  \label{eq:sel-pi}\\
  \mathcal{L}^{\mathrm{sel}}_{\mathrm{val}}(g) &= \textstyle\sum_i g_i\,(\yhat_i - y_i)^2 \,\big/\, \sum_i g_i .
  \label{eq:sel-val}
\end{align}
Relative to~\eqref{eq:qd}, every sample is weighted by the gate, and the coverage penalty is scaled by the \emph{selected} count $\sqrt{S_g}$ rather than the batch size, because that is the amount of data backing the $\picp_g$ estimate (Appendix~\ref{app:loss}). The auxiliary triple is trained with the plain IPIV loss on all samples, $\mathcal{L}_{\mathrm{aux}} = \beta_h\,\mathcal{L}^{(h)}_{\mathrm{PI}} + (1-\beta_h)\,\mathcal{L}^{(h)}_{v}$; a full triple, rather than one auxiliary regressor, keeps interval-relevant as well as point-relevant features alive for the inputs the gates reject.

\subsection{Two failure modes, two gates}
\label{sec:dual}
If a single gate weights both losses, it receives gradient from two sources that reward different behavior: the coverage penalty pushes $g$ down where the interval is likely to miss, the value loss where the point estimate is likely to be far from the target. By the coupling of \S\ref{sec:background} the second signal is bounded by width on captured inputs and the first is not, so one scalar must encode two different predictors, with the balance set implicitly by $\lambda_{\mathrm{PI}}$ rather than by the decision-maker. \spp{} therefore uses two heads, each weighting the loss whose failures it predicts and each with its own budget $\phi^{\mathrm{PI}}_{\mathrm{t}}, \phi^{\mathrm{val}}_{\mathrm{t}}$. An input is accepted only if \emph{both} gates admit it,
\begin{equation}
  \text{accept } x \iff \gpi(x) \ge \tau_{\mathrm{PI}} \;\wedge\; \gval(x) \ge \tau_{\mathrm{val}},
  \label{eq:product}
\end{equation}
with the two thresholds calibrated independently on validation data, each to its own budget. Figure~\ref{fig:arch} (right) shows the three cases this separates: a wide interval that is correct, which both gates accept and a width rule rejects; a narrow interval that misses, which only $\gpi$ rejects; and a captured target with the value estimate at the wrong end, which $\gval$ rejects.

Two gates with two budgets leave the quantity that actually matters, the \emph{joint} acceptance rate, uncontrolled. Since $(1-\gpi)(1-\gval) \ge 0$ we have $\gpi\gval \ge \gpi + \gval - 1$ pointwise, so the joint rate obeys the Fr\'echet bound
\begin{equation}
  \hat{\phi}_{\mathrm{joint}} \;=\; \tfrac{1}{n}\textstyle\sum_i \gpi(x_i)\,\gval(x_i) \;\ge\; \phi^{\mathrm{PI}}_{\mathrm{t}} + \phi^{\mathrm{val}}_{\mathrm{t}} - 1 .
  \label{eq:frechet}
\end{equation}
We make \eqref{eq:frechet} a \emph{trained} constraint: a quadratic penalty fires whenever the gates drift so far apart that their joint mass falls below the floor their budgets imply,
\begin{equation}
  \mathcal{L}_{\mathrm{fr}} = \lambda_{\mathrm{fr}}\,\Psi\big(\phi^{\mathrm{PI}}_{\mathrm{t}} + \phi^{\mathrm{val}}_{\mathrm{t}} - 1 - \hat{\phi}_{\mathrm{joint}}\big).
  \label{eq:frechet-loss}
\end{equation}
Without it the two heads are free to reject disjoint sets, and the intersection collapses well below either budget. With $\mathcal{L}_{\gpi} = \beta_{\mathrm{PI}}\,\mathcal{L}^{\mathrm{sel}}_{\mathrm{PI}}(\gpi) + (1-\beta_{\mathrm{PI}})\,\mathcal{L}_{\mathrm{fr}}$, the full objective mixes the two selective losses as IPIV mixes its losses, and the result with the auxiliary loss as SelectiveNet does:
\begin{equation}
  \mathcal{L}_{\mathrm{2G}} = \gamma\Big[\beta\,\mathcal{L}_{\gpi} + (1-\beta)\,\mathcal{L}^{\mathrm{sel}}_{\mathrm{val}}(\gval)\Big] + (1-\gamma)\,\mathcal{L}_{\mathrm{aux}} .
  \label{eq:dual-loss}
\end{equation}
We fix $\gamma = \beta = \beta_h = 0.5$, the parents' defaults; $\lambda_{\mathrm{fr}}$ and $\beta_{\mathrm{PI}}$, which have no counterpart in either parent, are tuned per dataset. Tying the gates ($\gpi \equiv \gval \equiv g$, $\lambda_{\mathrm{fr}} = 0$) gives the \emph{single-gate} variant \sppone{}, which we keep as an ablation; the full model is \spptwo{}. The budget pair (e.g.\ $0.85$ each guarantees a joint rate of at least $0.70$) also exposes a \emph{frontier} that trades selective interval quality against selective point accuracy at a fixed joint rate (Appendix~\ref{app:frontier}), which is why we do not tie the two budgets to one joint target by formula.

\subsection{Training and inference}
\label{sec:train}
All heads are trained end to end with Adam; depth, epochs, $\lambda_{\mathrm{PI}}$ and initialization are inherited unchanged from IPIV's published per-dataset configuration, except the learning rate, which we cap at $5\times10^{-3}$ because IPIV's larger values destabilize the multi-term objective. All penalties are active from the first epoch: a warm start that delays the coverage penalty lets the two bounds cross and never recover (Appendix~\ref{app:loss}). One model is trained per target acceptance rate; at inference the thresholds are set on a validation split to accept the budgeted fraction of inputs, and every baseline's threshold is calibrated on the same split. The model returns $\big(\yhat(x), [L(x), U(x)]\big)$ on accepted inputs, from a single forward pass, and abstains otherwise.

\section{Evaluation Design}
\label{sec:setup}

\begin{table}[t]
\caption{\textbf{Evaluation design.} \spp{} has no like-for-like competitor, so each question is asked on the home ground of the field it comes from, with that field's comparator and metric.}
\label{tab:design}
\centering
\footnotesize
\setlength{\tabcolsep}{3.5pt}
\begin{tabular}{@{}l p{4.35cm} p{3.9cm} p{3.6cm}@{}}
\toprule
 & Question (origin) & Comparator & Metric \\
\midrule
Q1 & Does learning to abstain damage the interval model? (\emph{PI field}) & IPIV, all inputs answered, IPIV's protocol and datasets & $\picp$, $\mpiw$, RMSE at $\phi = 1$ \\
Q2 & Is it a competitive selective regressor? (\emph{selective-prediction field}) & SelectiveNet-regression, its own grid and settings & selective RMSE / MAE vs.\ $\phi$ \\
Q3 & Given an interval model, does a learned gate beat the model's own uncertainty? (\emph{new}) & same predictor, ranked by gate / by width $U{-}L$ / by an oracle; a deep ensemble ranked by member disagreement & selective interval score and RMSE vs.\ $\phi$; selection efficiency \\
\bottomrule
\end{tabular}
\end{table}

\paragraph{Rationale.}
\spp{} produces an output---interval, value and abstention decision---that no existing method produces, so no single comparison is like-for-like. PI methods answer every query and are scored on the $(\picp,\mpiw)$ pair; selective methods abstain and are scored on risk--coverage curves of a point loss. Neither convention transfers on its own. Under selection the $(\picp,\mpiw)$ pair is no longer a fair summary, because a selector can buy narrow accepted intervals by keeping the misses, or perfect coverage by keeping only easy inputs. A point-loss risk--coverage curve ignores the interval altogether and mechanically favors any width-based selector (\S\ref{sec:background}). We therefore ask three questions, each on the home ground of the field it comes from (Table~\ref{tab:design}). Q1 and Q2 are the checks each parent field would make first, under that field's protocol. Q3 is the contribution, and it needs two instruments neither field supplies: a \emph{same-predictor} comparison, which ranks the test points of one trained model by the gate, by interval width and by an oracle, so that differences are attributable to the selector and not to the base model; and a \emph{proper} interval score, which charges for a miss rather than only for width and yields one number per (dataset, $\phi$) cell. 
\paragraph{Datasets.}
We use 20 regression tasks (Appendix~\ref{app:data}). \emph{(A) The UCI PI suite}, the ten datasets of QD and IPIV under the protocol of \citet{hernandez2015probabilistic}, and \emph{(B) IPIV's two image tasks}, RSNA bone age~\citep{halabi2019rsna} and IMDB-WIKI apparent age~\citep{rothe2018deep}, are the PI field's home ground (Q1). Several UCI sets are too small to evaluate selection (at $\phi = 0.3$ Yacht accepts 9 test points per split), so we add \emph{(C) eight larger OpenML regression tasks}~\citep{fischer2023openml,grinsztajn2022tree} with $20$k--$582$k samples, for statistical power. Bone age has a single 200-image test split on which validation-calibrated thresholds transfer poorly, so we treat it as illustrative; IMDB-WIKI (5-fold CV, ${\sim}172$k images) is the reliable image benchmark.

\paragraph{Baselines.}
All baselines share splits and seeds, and every threshold is calibrated on validation data.
\begin{itemize}[leftmargin=1.1em,nosep]
  \item \emph{IPIV + width threshold}, the post-hoc rule the central claim is stated against. We evaluate it in two forms: \emph{same-predictor} (ranking \spp{}'s own intervals by width; used for Q3 and for selection efficiency) and \emph{cross-pipeline} (a separately trained IPIV ranked by its own width; Appendix~\ref{app:full}).
  \item \emph{Deep-ensemble disagreement}, the strongest post-hoc rival and the textbook epistemic signal: $M{=}5$ IPIV models per split, combined with the aggregation IPIV's own paper reports (adapted from \citealp{lakshminarayanan2017simple}) and ranked by the variance of the members' point predictions (\emph{point disagreement}) or the summed spread of their bounds (\emph{bound disagreement}), each against the ensemble's own width threshold and oracle. The comparison is cross-pipeline: one \spp{} model against a five-model ensemble, at a fifth of the training and inference cost.
  \item \emph{SelectiveNet-regression}, learned selection on a point regressor; it produces no interval and appears only in point-error comparisons (Q2).
  \item \emph{Reference points}: IPIV at full acceptance, and an \emph{oracle} that ranks by realized loss and upper-bounds any selector on the same predictor. The non-selective methods of the PI literature (QD, QD+, deep ensembles) are compared at full acceptance in Appendix~\ref{app:published}.
\end{itemize}
Two asymmetries should be stated: \spptwo{}'s budgets are tuned by search whereas SelectiveNet runs at its published settings, so we report the \emph{median} search trial, not the best; and SelectiveNet uses its own backbone (Appendix~\ref{app:hyper}).

\paragraph{Protocol and metrics.}
One \spp{} model is trained per target acceptance rate on the grid $\phi \in \{0.30, 0.40, \ldots, 0.70, 0.75, \ldots, 0.95\}$, which contains SelectiveNet's regression grid, with $1-\alpha = 0.95$ throughout; the post-hoc selectors are swept. Tabular results average 5 random 90/10 splits (the provider's split for Year-MSD). Every metric is a selective risk~\eqref{eq:selrisk} as a function of $\phi$. The headline is the \textbf{selective interval score}, the accepted-set average of the Winkler score~\citep{winkler1972decision},
\begin{equation}
  \IS_\alpha(L, U; y) \;=\; (U - L) \;+\; \tfrac{2}{\alpha}\,(L - y)\,\ind[y < L] \;+\; \tfrac{2}{\alpha}\,(y - U)\,\ind[y > U],
  \label{eq:winkler}
\end{equation}
a strictly proper scoring rule for interval forecasts~\citep{gneiting2007strictly}. We also report its components (selective $\picp$, $\mpiw$), selective RMSE/MAE, and the \emph{value-position error} $|v_i - v_i^\star|$ on captured inputs, which $\gval$ should select on. Curves are summarized by \textbf{AURC}, the normalized area under the selective-risk curve over $\phi \in [0.3,1]$, and by \textbf{selection efficiency}
$\mathrm{SE} = (\aurc_{\text{width}} - \aurc_{\text{gate}})/(\aurc_{\text{width}} - \aurc_{\text{oracle}})$,
the fraction of the headroom between the same-predictor width threshold and the oracle that the gate captures; SE is the excess-AURC of \citet{geifman2018bias} made unit-free so that it can be aggregated across datasets. Gate diagnostics (AUROC for each failure mode, Spearman $\rho$ with width) do not depend on the base predictor. The unit of statistical analysis is the per-dataset AURC, compared with a Wilcoxon signed-rank test~\citep{demvsar2006statistical} and always accompanied by win/loss counts (Appendices~\ref{app:metrics} and \ref{app:stats}).

\section{Results}
\label{sec:results}

\begin{table}[t]
\caption{\textbf{Per-dataset results}, 18 tabular datasets, $1-\alpha = 0.95$, \spptwo{} median search trial throughout. \emph{Left (Q2):} RMSE of IPIV answering every query; RMSE reduction of \spptwo{} at two realized acceptance levels; SelectiveNet-regression vs.\ \spptwo{} at $\phi = 0.7$ ($\Delta > 0$ favors \spp{}). \emph{Right (Q3):} learned gate vs.\ width threshold on the \emph{same} predictor at the realized acceptance nearest $0.7$; lower $\IS_\alpha$ is better.}
\label{tab:main}
\centering
\scriptsize
\setlength{\tabcolsep}{3.2pt}
\resizebox{\textwidth}{!}{%
\begin{tabular}{@{}l r rr rrr cc cc@{}}
\toprule
 & IPIV & \multicolumn{2}{c}{\spptwo{} RMSE cut vs.\ IPIV} & \multicolumn{3}{c}{sel.\ RMSE at $\phi = 0.7$} & \multicolumn{2}{c}{\spptwo{} sel.\ $\picp$ at $\phi{\approx}0.7$} & \multicolumn{2}{c}{\spptwo{} sel.\ $\IS_\alpha$ at $\phi{\approx}0.7$} \\
\cmidrule(lr){3-4}\cmidrule(lr){5-7}\cmidrule(lr){8-9}\cmidrule(l){10-11}
Dataset & $\phi{=}1$ & $\phi{\approx}0.55$ & $\phi{\approx}0.75$ & Sel.Net & \spptwo{} & $\Delta$ & gate & width & gate & width \\
\midrule
yacht              & 1.04   & 63.3\% & 56.1\% & 1.62   & \textbf{0.35}   & $+78.3\%$ & \textbf{0.975} & 0.893 & \textbf{0.10} & 0.45 \\
boston             & 2.64   & 10.5\% & 7.5\%  & 2.42   & \textbf{2.39}   & $+1.3\%$  & \textbf{0.934} & 0.869 & \textbf{1.72} & 2.16 \\
energy             & 1.63   & 59.4\% & 50.5\% & 1.17   & \textbf{0.82}   & $+30.2\%$ & \textbf{0.945} & 0.855 & \textbf{0.45} & 1.41 \\
concrete           & 6.17   & 33.5\% & 22.2\% & 4.66   & \textbf{4.59}   & $+1.5\%$  & \textbf{0.850} & 0.809 & \textbf{1.66} & 2.03 \\
wine               & 0.66   & 16.2\% & 10.1\% & 0.56   & 0.56            & $0.0\%$   & \textbf{0.812} & 0.701 & \textbf{5.39} & 9.29 \\
kin8nm             & 0.08   & 11.0\% & 1.5\%  & \textbf{0.07}   & 0.08   & $-15.5\%$ & \textbf{0.880} & 0.786 & \textbf{1.68} & 3.78 \\
power-plant        & 4.10   & 7.5\%  & 3.2\%  & \textbf{3.86}   & 3.92   & $-1.5\%$  & \textbf{0.937} & 0.924 & \textbf{1.14} & 1.20 \\
naval              & 0.0010 & 15.9\% & 3.0\%  & \textbf{0.0006} & 0.0011 & $-83.6\%$ & \textbf{0.956} & 0.812 & \textbf{0.30} & 2.47 \\
protein            & 4.41   & 28.3\% & 14.4\% & 3.93   & \textbf{3.56}   & $+9.4\%$  & \textbf{0.947} & 0.861 & \textbf{2.87} & 5.19 \\
Year-MSD           & 9.01   & 43.1\% & 28.9\% & 6.13   & \textbf{5.96}   & $+2.8\%$  & \textbf{0.932} & 0.916 & \textbf{3.35} & 4.38 \\
superconduct       & 13.05  & 56.9\% & 39.5\% & 8.63   & \textbf{7.40}   & $+14.3\%$ & \textbf{0.941} & 0.887 & \textbf{1.11} & 2.22 \\
sarcos             & 3.23   & 51.5\% & 35.3\% & 2.40   & \textbf{1.91}   & $+20.3\%$ & \textbf{0.955} & 0.929 & \textbf{0.46} & 0.60 \\
diamonds           & 563.4  & 73.3\% & 56.6\% & 237.6  & \textbf{232.5}  & $+2.2\%$  & \textbf{0.973} & 0.890 & \textbf{0.26} & 2.28 \\
video\_transcoding & 2.83   & 87.5\% & 80.1\% & 0.79   & \textbf{0.50}   & $+37.4\%$ & \textbf{0.982} & 0.974 & \textbf{0.16} & 0.23 \\
medical\_charges   & 0.08   & 30.9\% & 19.6\% & 0.07   & \textbf{0.06}   & $+10.2\%$ & \textbf{0.961} & 0.955 & \textbf{0.60} & 0.66 \\
black\_friday      & 0.39   & 8.1\%  & 4.9\%  & 0.38   & \textbf{0.37}   & $+3.5\%$  & \textbf{0.950} & 0.856 & \textbf{3.55} & 9.25 \\
particulate-matter & 0.39   & 29.7\% & 19.1\% & 0.32   & \textbf{0.30}   & $+4.1\%$  & \textbf{0.943} & 0.935 & \textbf{2.32} & 2.51 \\
nyc-taxi-green     & 0.41   & 31.7\% & 21.8\% & 0.34   & \textbf{0.31}   & $+9.6\%$  & \textbf{0.958} & 0.887 & \textbf{2.98} & 7.20 \\
\midrule
median [IQR]       &        & 31\% [16--56] & 21\% [8--38] &  &      & $+3.8\%$ [$+1.4$, $+13.3$] & \textbf{0.946} & 0.887 & \textbf{1.40} & 2.25 \\
wins / ties / losses &      & 18/0/0 & 18/0/0 & \multicolumn{3}{c}{14 / 2 / 2 for \spptwo{}} & \multicolumn{2}{c}{18/0/0} & \multicolumn{2}{c}{18/0/0} \\
\bottomrule
\end{tabular}%
}
\end{table}

\paragraph{Q1: the gates are nearly free; the selective objective is not.}
At $\phi = 1$ the gates are inactive, so \spp{} is an IPIV model trained with the selective objective and carrying extra heads. Across the 18 tabular datasets it deviates from plain IPIV by a median $-0.013$ in $\picp$ (a small loss of coverage), $-6.9\%$ in $\mpiw$ (narrower intervals, an improvement) and $+4.4\%$ in RMSE (a loss); per-dataset values are in Table~\ref{tab:perdataset}. Two effects are mixed here; setting the acceptance target to $1$ separates them, because the selective losses then reduce to IPIV's own and only the architecture differs: \textbf{the extra heads alone cost a median $+0.6\%$ in RMSE, $-0.008$ in $\picp$ and $-4.2\%$ in $\mpiw$, and improve on IPIV on 7 of 18 datasets.} The rest of the RMSE deviation comes from the selective objective, which optimizes the model for the inputs it will accept; that is the effect the method exists to exploit, and a price only if one trains to select and then declines to. The architecture costs more than $10\%$ in RMSE on four datasets (kin8nm, naval, video\_transcoding, black\_friday), with no relation to heteroscedasticity ($\rho = -0.22$, $p = 0.39$). The answer to Q1: adding the abstention machinery leaves the interval model essentially intact, with about one point of coverage lost, narrower intervals, and a few percent of RMSE lost mostly to the selective objective rather than to the added heads.

\paragraph{Q2: intervals are usually not paid for in point accuracy.}
Abstaining pays in absolute terms: at a realized acceptance of ${\sim}0.55$, \spptwo{} cuts RMSE relative to IPIV on 18 of 18 datasets, by a median $31\%$ (Table~\ref{tab:main}) and by up to $88\%$ where error is concentrated in a minority of inputs (video\_transcoding, diamonds); black\_friday and naval gain least and are the weak cases of every later analysis. The informative comparison is with SelectiveNet-regression, a method with no interval to maintain. At $\phi = 0.7$ the median \spptwo{} trial has the lower selective RMSE on 14 of 18 tabular datasets, ties on two (wine, power-plant) and loses on two (kin8nm, naval), both near-deterministic tasks with little uncertainty for an interval method to characterize. The image tasks are less favorable (Appendix~\ref{app:images}). On IMDB-WIKI \textbf{SelectiveNet wins wherever the two overlap}, by $1.3$--$1.5$ years of MAE ($6.29$ vs.\ $7.59$ at $\phi \in [0.7,0.8)$); on bone age the median \spptwo{} trial is slightly behind ($16.2$ vs.\ $15.9$ months at $[0.7,0.8)$), though the best trial wins every bucket. We attribute the IMDB-WIKI result to headroom: IPIV is already tight there, little error sits in a rejectable minority, and a single-purpose objective extracts what remains more efficiently than our joint one (\S\ref{sec:analysis}). Part of any point-error advantage is, moreover, the structural coupling of \S\ref{sec:background}, which is why Q3, not Q2, is our evidence that the gates learn something.

\begin{figure}[t]
\begin{center}
\includegraphics[width=0.305\textwidth]{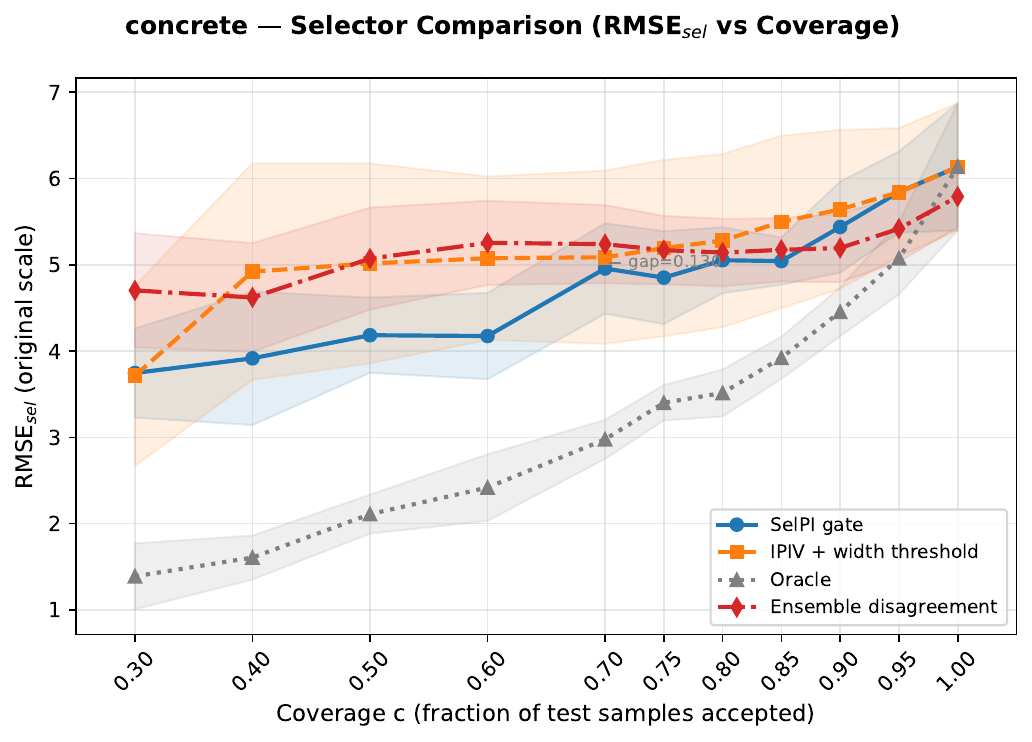}\hfill
\includegraphics[width=0.305\textwidth]{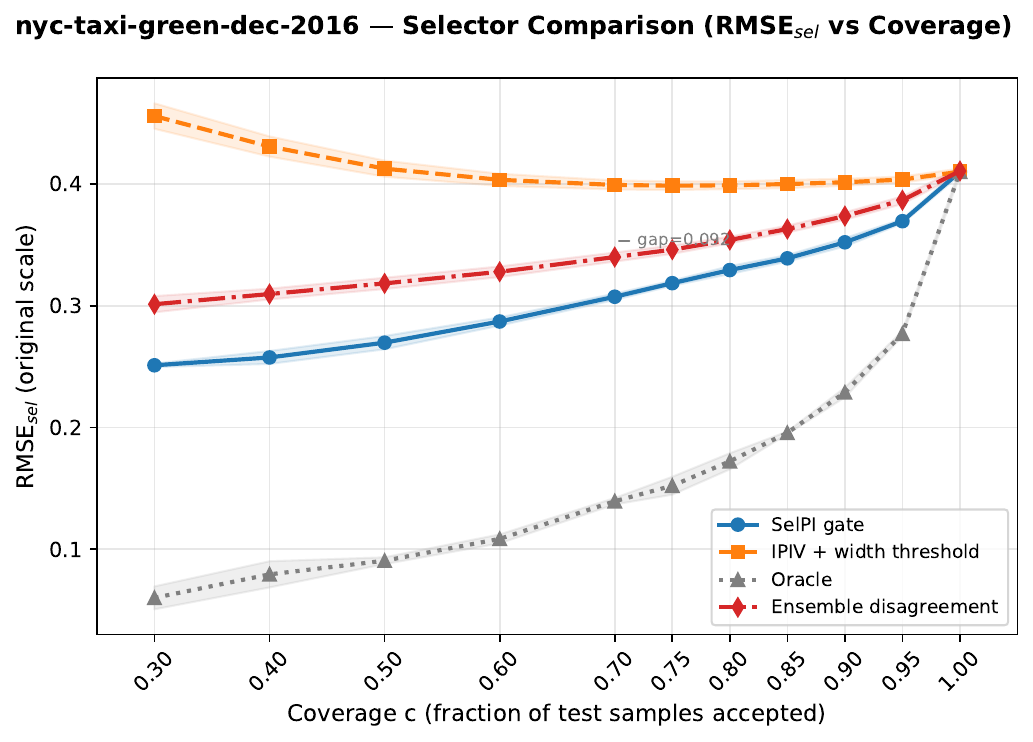}\hfill
\includegraphics[width=0.305\textwidth]{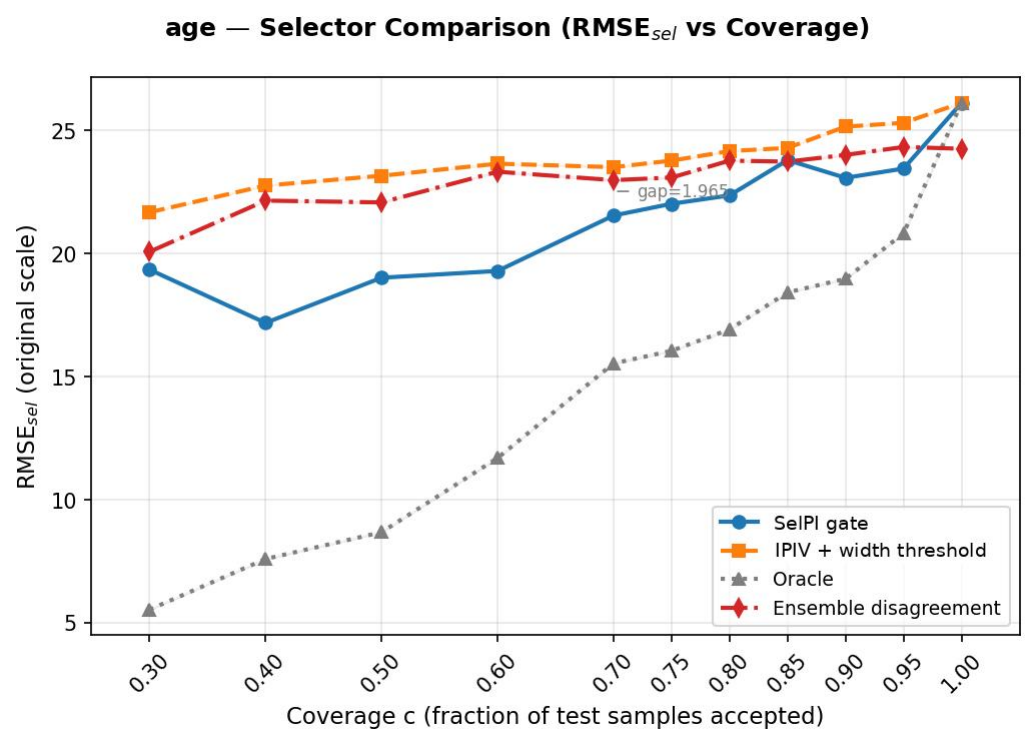}
\end{center}
\caption{\textbf{Q3: selective RMSE vs.\ acceptance rate on a fixed predictor} (single-gate \sppone{}, one model per acceptance rate) for Concrete, NYC Taxi and bone age (illustrative only, \S\ref{sec:setup}). Curves: learned gate, width threshold, oracle, and deep-ensemble point disagreement on its own five-model predictor; bands: $\pm 1$ s.d.\ over runs; lower is better.}
\label{fig:curves}
\end{figure}

\paragraph{Q3: a learned gate beats width thresholding, most clearly on interval quality.}\label{sec:main-claim}
Figure~\ref{fig:curves} shows three datasets; per-dataset values for all 20 are in Figure~\ref{fig:se} and Table~\ref{tab:ensemble}. With the base predictor held fixed, the learned gate attains a lower selective-RMSE AURC than the width threshold on all 20 datasets (Wilcoxon $p = 2\times10^{-6}$), with a median selection efficiency of $0.59$ ($0.65$ on the 18 tabular datasets, IQR $0.48$--$0.82$). On the interval score the margin is far larger: 20 of 20, median SE $\mathbf{0.87}$ (IQR $0.74$--$0.96$). The smallest gain is on IMDB-WIKI (SE $0.08$: interval score $51.95$ against $52.99$ for width, oracle $38.20$), consistent with its tight, near-homoscedastic predictor (\S\ref{sec:analysis}). An oracle gap remains everywhere; the gate does not close it, and is closest on the most heteroscedastic datasets.

The gap between the two metrics is informative: a width threshold is a tolerable point-error selector but a poor interval selector. Rejecting wide intervals removes the inputs the model \emph{covered}, so the accepted set retains the misses, and each miss costs $2/\alpha = 40$ times its distance outside the interval. Decomposing $\IS_\alpha$ at $\phi \approx 0.7$ confirms the mechanism (right block of Table~\ref{tab:main}): the gate accepts intervals \emph{wider} than the width threshold's by a median $9.3\%$ and still wins, because it cuts the miss-penalty term by a median $71.0\%$ (miss rate $0.054$ against $0.113$). Width buys sharpness on the accepted set and pays in misses; the gate makes the opposite trade, and under a proper scoring rule that is the better one.

\begin{table}[tb]
\caption{\textbf{What each signal sees, and how well it selects.} AUROC for coverage failure (cov.) and value failure (val.; captured, but error above the captured-set median), and median SE against each selector's \emph{own} predictor's width threshold and oracle (${<}0$: worse than width). The SE on the $\gpi$/$\gval$ rows is that of the Q3 gate (\sppone{}); Pt./Bd.\ disagr.\ is the deep-ensemble baseline on its own five-model predictor. Medians over 18 tabular datasets.}
\label{tab:auroc}
\centering
\footnotesize
\setlength{\tabcolsep}{6pt}
\begin{tabular}{@{}l cc cc@{}}
\toprule
 & \multicolumn{2}{c}{AUROC} & \multicolumn{2}{c}{selection efficiency} \\
\cmidrule(lr){2-3}\cmidrule(l){4-5}
Signal & cov.$\uparrow$ & val.$\uparrow$ & sel.\ RMSE & $\IS_\alpha$ \\
\midrule
Width $U - L$   & 0.378          & \textbf{0.717} & \multicolumn{2}{c}{0 (reference)} \\
$\gpi$          & \textbf{0.790} & 0.633 & \multirow{2}{*}{$\mathbf{0.59}$} & \multirow{2}{*}{$\mathbf{0.87}$} \\
$\gval$         & 0.710          & 0.647 & & \\
Pt.\ disagr.    & 0.584          & 0.606 & $-0.34$ & $-0.62$ \\
Bd.\ disagr.    & 0.591          & 0.581 & $-0.37$ & $-0.55$ \\
\bottomrule
\end{tabular}
\end{table}
\paragraph{Q3, continued: the gate also beats ensemble disagreement.}
Width is not the only post-hoc alternative. The deep-ensemble baseline of \S\ref{sec:setup} ranks a five-model IPIV ensemble by point or bound disagreement across members, the textbook epistemic signal, and Table~\ref{tab:auroc} (right columns) compares it with the gate, each selector measured against its own predictor's width threshold and oracle. Comparing selection efficiency across the two pipelines, the gate wins on every dataset against both signals: 20 of 20 on selective RMSE (Wilcoxon $p = 2\times10^{-6}$) and 19 of 19 on the interval score ($p = 4\times10^{-6}$). The disagreement selectors mostly fail to beat even their own ensemble's width threshold: on selective RMSE, point disagreement loses to width on 18 of 20 datasets and bound disagreement on 18 of 20; on the interval score, 13 of 19 and 14 of 19 respectively (yacht excluded). The left columns of Table~\ref{tab:auroc} show why: both signals sit well below $\gpi$ on coverage failure. Per-dataset values are in Appendix~\ref{app:ensemble}. Variance across independently trained networks does not substitute for a gate trained jointly with the predictor. (Yacht is excluded from the ensemble's $\IS_\alpha$ column, where its SE is undefined; Appendix~\ref{app:metrics}.)

\paragraph{Two gates vs.\ one.}
At matched acceptance \spptwo{} beats \sppone{} in selective RMSE on all 18 tabular datasets while producing \emph{wider} intervals on 16 of them---the opposite of what an ``it just picks easy points'' explanation predicts---and it reduces the value-position error $|v - v^\star|$ on its accepted set on 18 of 18 (median $10.6\%$), so $\gval$ selects on the quantity it is trained for rather than on width by proxy. The two-budget frontier (Appendix~\ref{app:frontier}) shows the mechanism: shifting budget between the gates at a fixed joint rate trades interval quality against point accuracy, which one scalar gate cannot express. On images the picture is weaker: \spptwo{} wins on bone age, but on IMDB-WIKI the single gate is better at three of five comparable acceptance rates.

\section{Analysis: What the Gates Learn That Width Does Not}
\label{sec:analysis}

\paragraph{Complementarity and specialization.}\label{sec:complementarity}
Treating ``did the interval miss?'' and ``was the value estimate misplaced?'' as binary labels, Table~\ref{tab:auroc} asks how well each signal ranks them. Width predicts \emph{value} failure well ($0.717$), as the coupling of \S\ref{sec:background} implies, but on \emph{coverage} failure it scores $0.378$---below chance, and therefore anti-predictive: wider intervals miss less often, so a reject-if-wide rule systematically discards the inputs whose intervals were correct. $\gpi$ reaches $0.790$ on that same event. The gates' rank correlation with width stays moderate (per-dataset means from $-0.77$ to $+0.21$): width is a valid aleatoric signal, but reproducing it would be learning nothing. Splitting rejections by which gate \emph{alone} is responsible confirms the specialization: $\gval$'s unique rejections are enriched for value failure on 17 of 18 datasets, and $\gpi$'s for coverage failure on 13 of the 17 where it rejects anything uniquely (Table~\ref{tab:speclift}).

\paragraph{When does learned selection help?}\label{sec:when}
We relate each dataset's selection efficiency to three task properties; with 20 datasets this is descriptive, not inferential (Appendix~\ref{app:when}). Selection efficiency rises with residual heteroscedasticity (Spearman $\rho = 0.54$, $p = 0.013$), as predicted. The deep-ensemble selectors of Table~\ref{tab:auroc} show the opposite pattern (point disagreement $\rho = -0.87$, bound disagreement $\rho = -0.83$, both $p < 10^{-4}$), so the relationship is specific to a gate trained jointly with the predictor. Target kurtosis and full-acceptance miscoverage show no usable relationship (Appendix~\ref{app:when}). The extremes fit: video\_transcoding and diamonds sit near $0.99$, while particulate-matter, power-plant and IMDB-WIKI---tight, near-homoscedastic predictors---sit below $0.16$, where a width threshold is already close to optimal.

\paragraph{Shift-like subpopulations.}\label{sec:shift}
Width is aleatoric and shift is epistemic, so shift is where a learned gate should most clearly beat width. Without retraining, we take for each dataset the covariate that best tracks realized error, treat its top quintile as a shifted tail, and ask how much each selector's abstention rate rises there at its $\phi = 0.7$ threshold. The gate's abstention rises by a median $+0.193$ against $+0.046$ for width, and the gate reacts more strongly on 15 of 18 datasets. The deep-ensemble selectors react more than width here too (point disagreement $+0.131$, bound disagreement $+0.165$, on the 15 datasets where the tail is defined for all selectors), but less than the gate: ensembling captures only part of the epistemic signal the gate exploits. The tail is chosen with knowledge of the error, so this is weaker evidence than an injected corruption, but the direction is clear (Appendix~\ref{app:shift}).

\paragraph{Ablations and cost.}\label{sec:ablations}
\emph{Joint training is essential} (Table~\ref{tab:ablations}, Appendix~\ref{app:ablations}): a gate trained on a frozen IPIV backbone is a median $5.6\%$ worse and loses on 15 of 18 datasets, by up to $54\%$; the gates shape the representation rather than merely reading it. \emph{The auxiliary triple matters}, though less ($+2.0\%$, worse on 14 of 18). \emph{The $\sqrt{S_g}$ scaling does not pay for itself empirically} ($+0.2\%$, 9 worse and 9 better); we retain it on principle (\S\ref{sec:single}).\label{sec:cost} Each gate adds $16h + 65$ parameters for hidden width $h$ ($44\%$ on a small tabular model, ${\sim}0.2\%$ on the image backbones); training takes a median $1.63\times$ IPIV's wall-clock per model. Inference cost is unchanged: one forward pass (Appendix~\ref{app:cost}).

\section{Conclusion}
\label{sec:conclusion}
A regression model can learn to predict, bound and abstain jointly. The extra heads cost almost nothing at full acceptance; the resulting selective regressor matches or beats SelectiveNet on most tabular tasks but not on IMDB-WIKI; and on a fixed predictor a learned gate beats width thresholding everywhere on a proper interval score, because width is anti-predictive of the misses that score charges for. It also beats deep-ensemble disagreement on every dataset. Two limitations remain. The accepted set carries no finite-sample coverage guarantee, since abstention breaks the exchangeability that split-conformal calibration relies on; calibrating \emph{conditional on acceptance} would restore one and make \spp{} the joint-training counterpart of selective conformal risk control~\citep{xu2025selective}. And our gates abstain on \emph{hard} inputs, whereas some applications care whether \emph{acting} is expected to lose; retargeting $\gval$ to a downstream utility is a natural extension.

\subsubsection*{AI use statement}
Large language model (LLM) tools were used solely to assist with the writing of this manuscript, specifically for language polishing, including improving grammar, clarity, phrasing, and overall readability. These tools were not used to generate research ideas, design or conduct experiments, analyze results, or produce scientific content. All AI-assisted edits were reviewed and verified by the authors, who take full responsibility for the content of this paper.

\subsubsection*{Reproducibility statement}
Everything needed to reproduce our results is specified in the appendices. Appendix~\ref{app:data} lists every dataset with its source, size, target statistics, preprocessing, and split protocol (five random 90/10 splits with fixed seeds for the tabular datasets, the provider's split for Year-MSD, five-fold cross-validation for IMDB-WIKI, and the fixed 200-image test set for bone age). Appendix~\ref{app:loss} states the training objective in full, including the hard and soft capture indicators, the $\sqrt{S_g}$ scaling, and the absence of a warm start. Appendix~\ref{app:hyper} gives the architecture and every hyperparameter of \spp{} and of each baseline, the values inherited unchanged from IPIV and SelectiveNet, the per-dataset search spaces and tuning budgets, and the rule by which the median search trial is reported. Appendix~\ref{app:metrics} defines every metric, the validation-based threshold calibration used for all selectors, and the selection-efficiency normalization; Appendix~\ref{app:stats} describes the statistical tests behind every reported $p$-value and win/loss count. The deep-ensemble baseline (number of members, aggregation, and ranking signals) is specified in \S\ref{sec:setup} and Appendix~\ref{app:ensemble}, and compute and hardware in Appendix~\ref{app:cost}. Full per-dataset results for every experiment summarized in the main text are reported in Appendices~\ref{app:images}--\ref{app:ensemble}.

\subsubsection*{Ethics statement}
This work uses only publicly available benchmark datasets under their original licenses: the UCI and OpenML regression tasks, the RSNA pediatric bone age challenge data (de-identified radiographs released for research), and IMDB-WIKI (publicly posted celebrity photographs with age labels). No new data involving human subjects were collected, and no attempt was made to identify individuals. Age estimation from faces and radiographs is used here as a benchmark for interval prediction with abstention, not as a deployable system; both datasets are known to be demographically unbalanced, and a selector trained on them may abstain at different rates across groups, which we did not audit. More broadly, the accepted set of \spp{} carries no finite-sample coverage guarantee (\S\ref{sec:conclusion}), so an abstention decision should be read as a learned signal of predictive difficulty, not as a certification of safety, and any high-stakes use would require domain-specific validation and human oversight. The authors declare no competing interests.

\FloatBarrier
\bibliography{sp_piven}
\bibliographystyle{iclr2027_conference}

\appendix
\section{Extended Related Work: Selection with Conformal Guarantees}
\label{app:related}
Selective conformal risk control~\citep{xu2025selective} selects post hoc and then calibrates prediction \emph{sets} for classification on the accepted subset; dual-threshold conformal abstention~\citep{kumar2025beyond} does the same for perception models, and adaptive conformal inference~\citep{gibbs2021adaptive} tracks coverage online under drift. Conformalized selective regression~\citep{sokol2026conformalized} goes the other way and uses a conformal score as the confidence function of a post-hoc selective regressor. All of these take either the predictor or the selector as given and add a calibration layer; we learn the selector jointly with the intervals, on regression intervals rather than label sets, and view a conformal wrapper calibrated conditional on acceptance as a natural extension (\S\ref{sec:conclusion}). SelectiveNet's selection machinery is loss-agnostic, but IPIV's heads are regression-specific; a classification analogue would target prediction sets and is left to future work.

\section{Loss Details}
\label{app:loss}
\paragraph{Hard and soft capture in the IPIV loss.}
In \eqref{eq:qd} the hard indicator $k_i$ has zero gradient almost everywhere, so the coverage term uses the softened capture $\tilde{k}_i = \sigma\big(s\,(U_i - y_i)\big)\,\sigma\big(s\,(y_i - L_i)\big)$ with $s = 160$. The two indicators are \emph{not} interchangeable across the two terms: the soft $\tilde{k}$ enters only the coverage penalty, while $\mpiw_{\mathrm{capt}}$ uses the hard $k$, so that an interval which misses its target contributes exactly zero to the width objective and is never encouraged to shrink further. The $\sqrt{n}$ factor is the batch size entering as a confidence weight: a larger sample makes the estimate of $\picp$ more trustworthy and a shortfall correspondingly more costly. \spp{} preserves this split exactly in \eqref{eq:gated-stats}: $\tilde{k}$ in $\picp_g$, $k$ in $\mpiw_g$.

\paragraph{Why $\sqrt{S_g}$ and not $\sqrt{n}$.}
When coverage is estimated only on the (softly) accepted subset, the amount of data backing the estimate is $S_g = \sum_i g_i$, not $n$. Keeping $\sqrt{n}$ over-weights the coverage constraint by a factor $\sqrt{n/S_g}$---$1.8\times$ at $\phit = 0.3$, and more early in training, when $g$ sits near $0.5$ everywhere. Empirically the choice makes little difference (Table~\ref{tab:ablations}); we keep $\sqrt{S_g}$ because it is the consistent generalization of \eqref{eq:qd}.

\paragraph{Why there is no warm start.}
A warm start that holds $\lambda_{\mathrm{PI}} = 0$ for the first ${\approx}20\%$ of epochs, letting the gates and the value head settle before interval pressure is applied, seems natural but does not work: with the coverage penalty off nothing stops the two bound heads from crossing, and a non-trivial fraction of runs end the warm-up with $U(x) < L(x)$ and never recover once the penalty is switched on. The coverage penalty is not a constraint on an already-valid interval---it is what makes the interval well ordered in the first place---so we train with all penalties active from the first epoch.

\section{Dataset Details}
\label{app:data}

\begin{table}[h]
\caption{The 20 datasets. $n$ and $d$ are read directly from the data files as loaded by our pipeline. ``Test/split'' is the held-out count per repetition under the protocol in the last column. Epochs are the per-dataset budgets inherited from IPIV's published configuration for the UCI sets and sized by timing probes for the larger ones. Every tabular dataset uses the same 10-point acceptance grid $\phi \in \{0.30, 0.40, 0.50, 0.60, 0.70, 0.75, 0.80, 0.85, 0.90, 0.95\}$, one model trained per point.}
\label{tab:datasets}
\begin{center}
\footnotesize
\setlength{\tabcolsep}{4pt}
\begin{tabular}{l r r r r r l}
\toprule
Dataset & $n$ & $d$ & Test/split & Epochs & Reps & Split protocol \\
\midrule
\multicolumn{7}{l}{\emph{(A) UCI PI suite} --- the standard benchmark of QD and IPIV} \\
Yacht        & 308     & 6  & 31     & 2000 & 5 & 90/10 random \\
Boston       & 506     & 13 & 51     & 300  & 5 & 90/10 random \\
Energy       & 768     & 8  & 77     & 1200 & 5 & 90/10 random \\
Concrete     & 1{,}030 & 8  & 103    & 800  & 5 & 90/10 random \\
Wine         & 1{,}599 & 11 & 160    & 1000 & 5 & 90/10 random \\
Kin8nm       & 8{,}192 & 8  & 819    & 800  & 5 & 90/10 random \\
Power-plant  & 9{,}568 & 4  & 957    & 500  & 5 & 90/10 random \\
Naval        & 11{,}934& 16 & 1{,}193& 1000 & 5 & 90/10 random \\
Protein      & 45{,}730& 9  & 4{,}573& 600  & 5 & 90/10 random \\
Year-MSD     & 515{,}345 & 90 & 51{,}630 & 300 & 1 & fixed provider split \\
\midrule
\multicolumn{7}{l}{\emph{(C) Larger tabular regression} --- added for statistical power at low $\phi$} \\
superconduct       & 21{,}263  & 79 & 2{,}126  & 500 & 5 & 90/10 random \\
sarcos             & 48{,}933  & 21 & 4{,}894  & 400 & 5 & 90/10 random \\
diamonds           & 53{,}940  & 26 & 5{,}394  & 400 & 5 & 90/10 random \\
video\_transcoding & 68{,}784  & 24 & 6{,}879  & 350 & 5 & 90/10 random \\
medical\_charges   & 163{,}065 & 3  & 16{,}307 & 200 & 5 & 90/10 random \\
black\_friday      & 166{,}821 & 23 & 16{,}682 & 200 & 5 & 90/10 random \\
particulate-matter & 394{,}299 & 26 & 39{,}430 & 120 & 5 & 90/10 random \\
nyc-taxi-green     & 581{,}835 & 9  & 58{,}184 & 100 & 5 & 90/10 random \\
\midrule
\multicolumn{7}{l}{\emph{(B) Image regression} --- IPIV's own two image tasks} \\
RSNA bone age (VGG-16) & --- & --- & 200 & 50 & 1 & fixed split \\
IMDB-WIKI (DenseNet)   & $\sim$171{,}852 & --- & $\sim$34{,}370 & 90\,/\,29 & 5 & 5-fold CV \\
\bottomrule
\end{tabular}
\end{center}
\end{table}

Inputs and targets are standardized to zero mean and unit variance; all interval widths and the $\picp$ figures below are therefore in standardized units, while point errors are reported on the original scale. $d$ counts features after one-hot expansion of categoricals for the group-C sets. Naval ships two regression targets and we keep the first, following the convention of the QD and IPIV evaluations. For IMDB-WIKI the two epoch budgets are $90$ for \spp{} and $29$ for IPIV and SelectiveNet-regression; the baselines converge within their shorter budget. Real per-epoch training curves confirm this is measured, not assumed: IPIV and SelectiveNet-regression plateau by epoch ${\sim}15$--$20$, while \spp{}'s harder joint objective --- interval, value, selection and auxiliary heads trained together --- was still improving at epoch 29, confirmed directly by retraining one fold to 90 epochs, where selective MAE fell from ${\sim}8.5$ to ${\sim}5.0$. The 90-epoch budget is not an inflated allowance chosen to favor \spp{}: it matches the original IPIV paper's own IMDB-WIKI training length exactly.

\section{Hyperparameters and Tuning Budget}
\label{app:hyper}

\paragraph{Method-level constants, fixed across all 20 datasets.}
$\gamma = \beta = \beta_h = 0.5$ (the convex-combination weights of \eqref{eq:dual-loss}, inherited from SelectiveNet's $\alpha$ and IPIV's $\beta$ respectively); nominal interval level $1-\alpha = 0.95$; softening factor $s = 160$ and acceptance-penalty weight $\lambda_{\mathrm{sel}} = 32$, taken unchanged from IPIV and SelectiveNet; gate hidden width $16$ with batch normalization; warm-start fraction $0$ (\S\ref{sec:train}); weight decay $0$, matching IPIV.

\begin{table}[h]
\caption{Per-dataset settings. Everything except the learning rate is inherited unchanged from IPIV's published configuration for that dataset. The learning rate is capped at $5\times10^{-3}$ because IPIV's larger values destabilize the multi-term selective objective (\S\ref{sec:train}); where the cap binds, the decay rate is raised to compensate.}
\label{tab:hyper}
\begin{center}
\footnotesize
\setlength{\tabcolsep}{5pt}
\begin{tabular}{l r r r r r}
\toprule
Dataset & Hidden & $\lambda_{\mathrm{PI}}$ & $\sigma_{\mathrm{init}}$ & Learning rate & Decay \\
\midrule
Boston       & 50  & 15.0 & 0.100 & $5\times10^{-3}$ & 0.990 \\
Concrete     & 50  & 15.0 & 0.200 & $5\times10^{-3}$ & 0.990 \\
Energy       & 50  & 17.0 & 0.350 & $5\times10^{-3}$ & 0.990 \\
Kin8nm       & 50  & 12.0 & 0.436 & $5\times10^{-3}$ & 0.990 \\
Naval        & 50  & 4.0  & 0.100 & $5\times10^{-3}$ & 0.990 \\
Power-plant  & 50  & 15.7 & 0.257 & $5\times10^{-3}$ & 0.990 \\
Wine         & 50  & 40.0 & 0.200 & $5\times10^{-3}$ & 0.990 \\
Yacht        & 50  & 6.0  & 0.200 & $5\times10^{-3}$ & 0.980 \\
Protein      & 100 & 40.0 & 0.200 & $2\times10^{-3}$ & 0.999 \\
Year-MSD and all 8 group-C sets & 100 & 40.0 & 0.200 & $2\times10^{-3}$ & 0.999 \\
\bottomrule
\end{tabular}
\end{center}
\end{table}

Batch size is $100$ for the UCI sets and $256$ for Year-MSD, medical\_charges, black\_friday, particulate-matter and nyc-taxi-green. Optimization is Adam with exponential decay throughout.

\paragraph{Dual-gate search.} The two acceptance budgets and the four weights with no counterpart in either parent method are tuned per dataset by Optuna (TPE), minimizing selective RMSE at a nominal joint acceptance rate of $0.7$, three seeds per trial. The search space is $\phi^{\mathrm{PI}}_{\mathrm{t}}, \phi^{\mathrm{val}}_{\mathrm{t}} \in [0.5, 0.9]$ uniform; $\lambda_{\mathrm{sel}}^{\mathrm{PI}}, \lambda_{\mathrm{sel}}^{\mathrm{val}} \in [8, 64]$ log-uniform; $\lambda_{\mathrm{fr}} \in [1, 32]$ log-uniform; $\beta_{\mathrm{PI}} \in [0, 1]$ uniform. Because the joint acceptance rate is measured rather than targeted (\S\ref{sec:dual}), the TPE population clusters in the middle of the range, so each dataset's main search is supplemented by dedicated low- and high-band batches that constrain the two budgets to $[0.50, 0.58]$ and $[0.82, 0.90]$ (and later $[0.90, 0.97]$) respectively. Trial counts after all batches are 149--287 per tabular dataset (180--287 on the ten UCI sets, 149--165 on the eight larger ones), 177 for bone age and 100 for IMDB-WIKI. Reported trials are filtered to those that are well calibrated ($\picp_{\mathrm{sel}} \ge 0.85$) and non-degenerate; results are then bucketed by realized joint acceptance rate in $0.1$-wide bins, and only buckets with at least 5 trials are reported.

\paragraph{SelectiveNet-regression baseline.} Its own canonical configuration, untuned: one hidden layer of 64 ReLU units with batch normalization, selection head of 16 units, learning rate $5\times10^{-4}$, 800 epochs, 5 runs, over its own published coverage grid. It brings its own backbone rather than sharing IPIV's, which is a real confound on the small nonlinear sets: its full-acceptance RMSE on Yacht is $4.03$ against IPIV's $1.04$, so a large part of the \spp{} margin there is backbone rather than selection.

\section{Metric Definitions}
\label{app:metrics}
Write $A_\phi = \{i : x_i \text{ accepted at rate } \phi\}$ for the accepted set, $S = |A_\phi|$, and $C = \{i \in A_\phi : k_i = 1\}$ for its captured subset. Every metric below is the instantiation of \eqref{eq:selrisk} with the stated per-sample loss.

\begin{itemize}[leftmargin=1.2em,itemsep=1pt,topsep=2pt]
  \item \textbf{Selective PICP}: $\picp(\phi) = \tfrac{1}{S}\sum_{i \in A_\phi} k_i$, reported as the deviation $|\picp(\phi) - (1-\alpha)|$, since over- and under-coverage are both failures.
  \item \textbf{Selective MPIW}: $\tfrac{1}{S}\sum_{i \in A_\phi}(U_i - L_i)$. Targets are standardized, so this is already in units of the target standard deviation.
  \item \textbf{Selective RMSE / MAE}: $\big(\tfrac{1}{S}\sum_{i \in A_\phi}(\yhat_i - y_i)^2\big)^{1/2}$ and $\tfrac{1}{S}\sum_{i \in A_\phi}|\yhat_i - y_i|$, on the original target scale. MAE is used for the two image tasks, following their convention; RMSE elsewhere.
  \item \textbf{Selective interval score}: $\tfrac{1}{S}\sum_{i \in A_\phi} \IS_\alpha(L_i, U_i; y_i)$ with $\IS_\alpha$ as in \eqref{eq:winkler}.
  \item \textbf{Value-position error}: $\tfrac{1}{|C|}\sum_{i \in C} |v_i - v_i^\star|$, $v_i^\star = (y_i - L_i)/(U_i - L_i)$, defined only on captured inputs since $v^\star \notin [0,1]$ otherwise; we report $|C|/S$ alongside it.
  \item \textbf{AURC}: the trapezoidal area of any of the above against $\phi$ over the grid $[0.3, 1]$, divided by $0.7$ so it reads as an average rather than an area. \textbf{E-AURC} subtracts the oracle computed on the same base predictor.
  \item \textbf{Selection efficiency} $\mathrm{SE} = (\aurc_{\text{width}} - \aurc_{\text{method}})/(\aurc_{\text{width}} - \aurc_{\text{oracle}})$: computed against the \emph{same-predictor} width threshold and oracle, never a separately trained IPIV's. $\mathrm{SE} = 1$ is the oracle, $\mathrm{SE} = 0$ the width threshold, and $\mathrm{SE} < 0$ a selector that is worse than a width threshold on its own predictor; SE is undefined when the width and oracle AURC coincide (yacht, for the ensemble's interval score). For the deep-ensemble selectors the ``own predictor'' is the five-model ensemble of \S\ref{sec:setup}, and width is its aggregated interval width, computed the same way on all three groups: IPIV's published aggregate (mean bounds widened by $z\,\sigma/\sqrt{M}$ across members). The disagreement signals' failure-mode AUROC (Table~\ref{tab:auroc}, and the corresponding columns of Table~\ref{tab:ensemble}) is reported on the 18 tabular datasets only, matching the scope of the diagnostic it extends; Appendix~\ref{app:ensemble}'s image rows report selection efficiency only, marked ``--'' in the AUROC columns.
\end{itemize}

Two events are used as binary labels for the gate diagnostics. A \emph{coverage failure} is $k_i = 0$: the interval missed. A \emph{value failure} is $k_i = 1$ together with $|\yhat_i - y_i|$ above the median absolute error among captured inputs: the interval was right but the value head was misplaced within it. The median is taken per dataset, so a value failure is by construction a within-dataset relative event, and roughly half of all captured inputs qualify. The pipeline standardizes targets before training, so widths are already in s.d.\ units, while point errors are converted back to the original scale --- width is therefore comparable across datasets and error is not, which is why selection efficiency, not raw RMSE, is the cross-dataset statistic.

\section{Image Regression Results}
\label{app:images}

\begin{table}[h]
\caption{\textbf{Full image-regression grids}, every trained point rather than the three buckets of Table~\ref{tab:images}. \emph{Top}: bone age, single gate, one run over 200 test images --- note the severe drift between target and realized acceptance rate, which is why Table~\ref{tab:images} is keyed by the realized rate. \emph{Middle}: IMDB-WIKI, single gate then the symmetric dual-gate grid, 5-fold CV at 90 epochs. \emph{Bottom}: SelectiveNet-regression on IMDB-WIKI over its own published coverage grid.}
\label{tab:imagegrids}
\begin{center}
\footnotesize
\setlength{\tabcolsep}{5pt}
\begin{tabular}{l rrrr}
\toprule
Target $\phi$ & Realized $\phi$ & $\picp_{\mathrm{sel}}$ & $\mpiw_{\mathrm{sel}}$ & MAE$_{\mathrm{sel}}$ \\
\midrule
\multicolumn{5}{l}{\emph{bone age, single gate --- IPIV at $\phi=1$: 0.915, 2.189, 20.71}} \\
0.30 & 0.000 & --- & --- & --- \\
0.40 & 0.070 & 0.929 & 1.589 & 10.901 \\
0.50 & 0.075 & 0.933 & 1.729 & 16.505 \\
0.60 & 0.190 & 1.000 & 2.063 & 14.099 \\
0.70 & 0.225 & 0.889 & 1.889 & 15.371 \\
0.75 & 0.480 & 0.917 & 1.775 & 16.381 \\
0.80 & 0.505 & 0.950 & 1.946 & 14.526 \\
0.85 & 0.625 & 0.904 & 1.869 & 16.198 \\
0.90 & 0.680 & 0.963 & 2.075 & 14.548 \\
0.95 & 0.865 & 0.879 & 1.933 & 18.027 \\
\midrule
\multicolumn{5}{l}{\emph{IMDB-WIKI, single gate, 90 epochs --- IPIV at $\phi=1$: 0.933, 2.835, 7.683}} \\
0.30 & 0.302 & 0.921 & 2.480 & 7.261 \\
0.40 & 0.401 & 0.943 & 2.854 & 7.691 \\
0.50 & 0.503 & 0.934 & 2.847 & 7.834 \\
0.60 & 0.602 & 0.935 & 2.857 & 7.826 \\
0.70 & 0.702 & 0.940 & 2.942 & 7.796 \\
0.75 & 0.750 & 0.942 & 3.019 & 7.975 \\
0.80 & 0.799 & 0.937 & 3.113 & 8.455 \\
0.85 & 0.851 & 0.938 & 3.137 & 8.416 \\
0.90 & 0.900 & 0.948 & 3.326 & 8.433 \\
0.95 & 0.959 & 0.926 & 3.259 & 9.113 \\
\midrule
\multicolumn{5}{l}{\emph{IMDB-WIKI, dual gate, symmetric budgets; second column is the realized \emph{joint} rate}} \\
0.5 & 0.490 & 0.942 & 2.884 & 7.641 \\
0.6 & 0.591 & 0.942 & 3.142 & 8.252 \\
0.7 & 0.694 & 0.945 & 3.217 & 8.446 \\
0.8 & 0.795 & 0.939 & 3.159 & 8.376 \\
0.9 & 0.898 & 0.932 & 3.201 & 8.841 \\
\midrule
\multicolumn{5}{l}{\emph{IMDB-WIKI, SelectiveNet-regression --- cols.\ are realized $\phi$, sel.\ MAE, full MAE, aux.\ MAE}} \\
0.70 & 0.700 & 6.294 & 7.257 & 7.256 \\
0.75 & 0.751 & 6.486 & 7.314 & 7.303 \\
0.80 & 0.799 & 6.506 & 7.202 & 7.204 \\
0.85 & 0.851 & 6.755 & 7.328 & 7.338 \\
0.90 & 0.901 & 6.847 & 7.242 & 7.242 \\
0.95 & 0.951 & 6.999 & 7.214 & 7.213 \\
\bottomrule
\end{tabular}
\end{center}
\end{table}
\FloatBarrier

\begin{figure}[h]
\begin{center}
\includegraphics[width=0.48\textwidth]{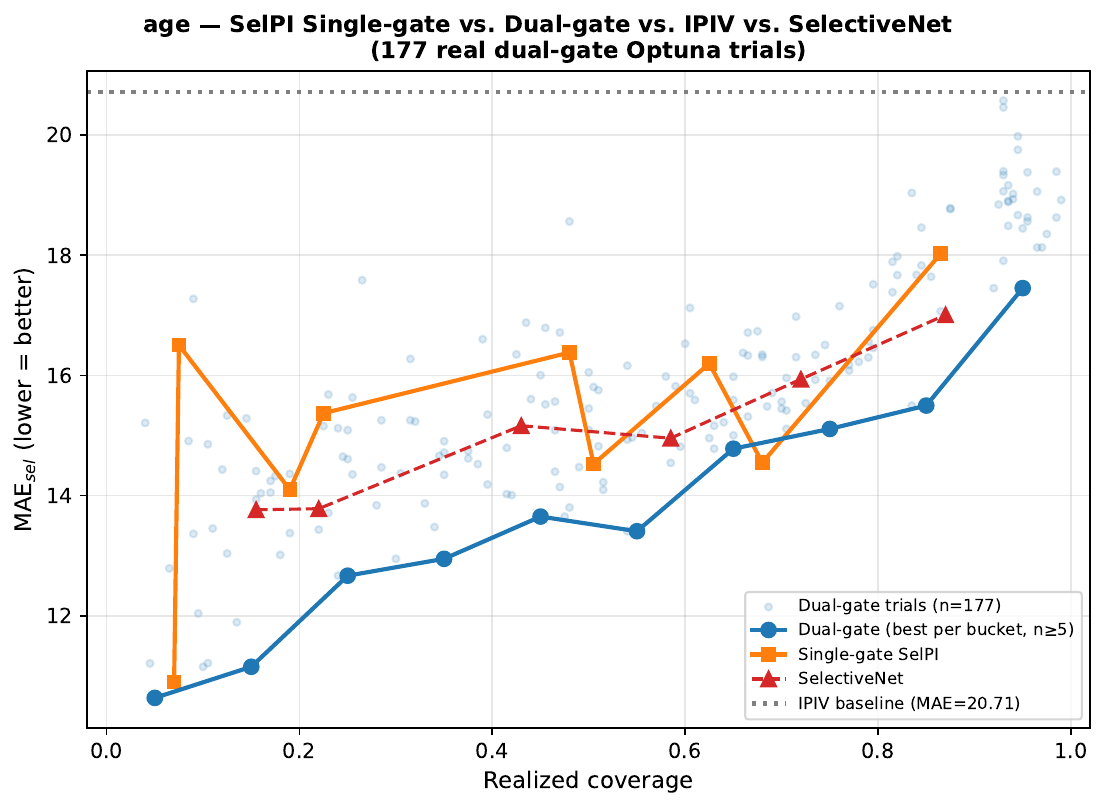}\hfill
\includegraphics[width=0.48\textwidth]{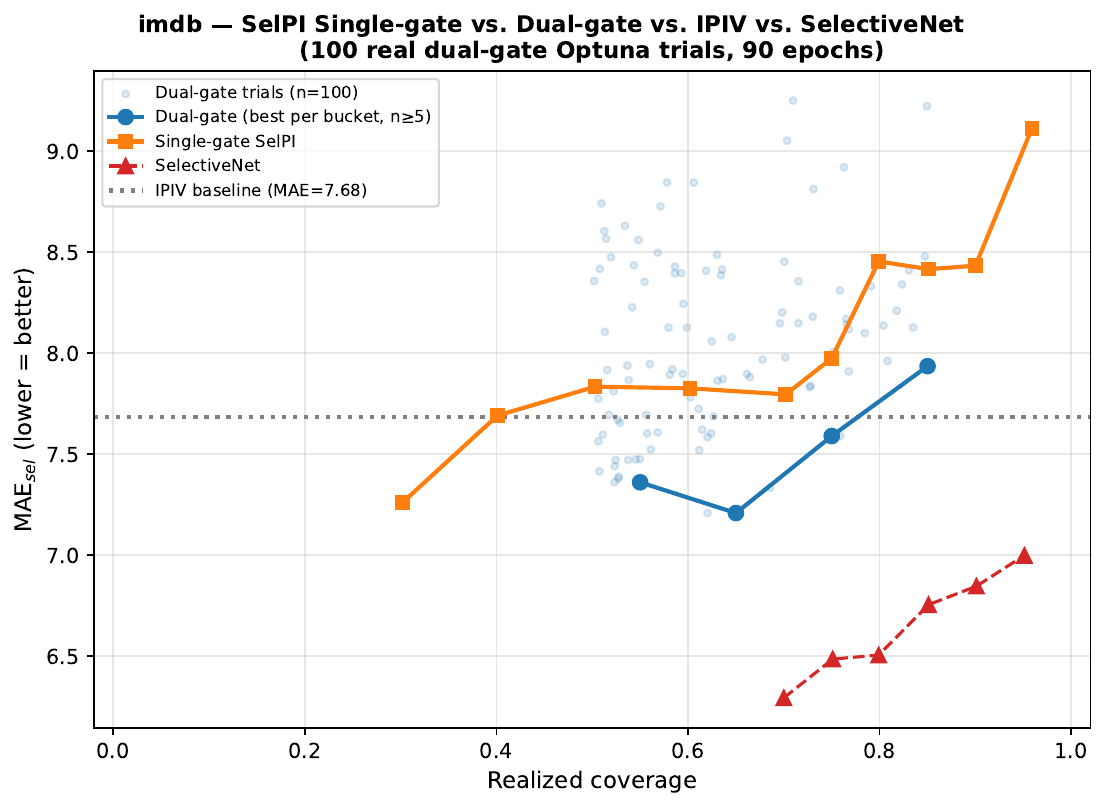}
\end{center}
\caption{\textbf{All four comparisons on the two image tasks}, at $1-\alpha = 0.95$: single-gate \spp{} (orange), dual-gate \spp{} (blue: every Optuna trial faintly, best-per-bucket as the line), plain IPIV at full acceptance (gray dotted), and SelectiveNet-regression (red). \emph{Left, bone age}: both \spp{} variants beat IPIV everywhere and the dual gate beats SelectiveNet at every comparable acceptance rate. \emph{Right, IMDB-WIKI}: the dual gate beats IPIV over the $0.5$--$0.8$ range but \textbf{SelectiveNet wins clearly and consistently}, by $0.7$--$1.6$ years of MAE. IMDB-WIKI is the largest dataset in the study and this is the paper's clearest negative result (\S\ref{sec:results}).}
\label{fig:images}
\end{figure}
\FloatBarrier

\begin{figure}[h]
\begin{center}
\includegraphics[width=0.49\textwidth]{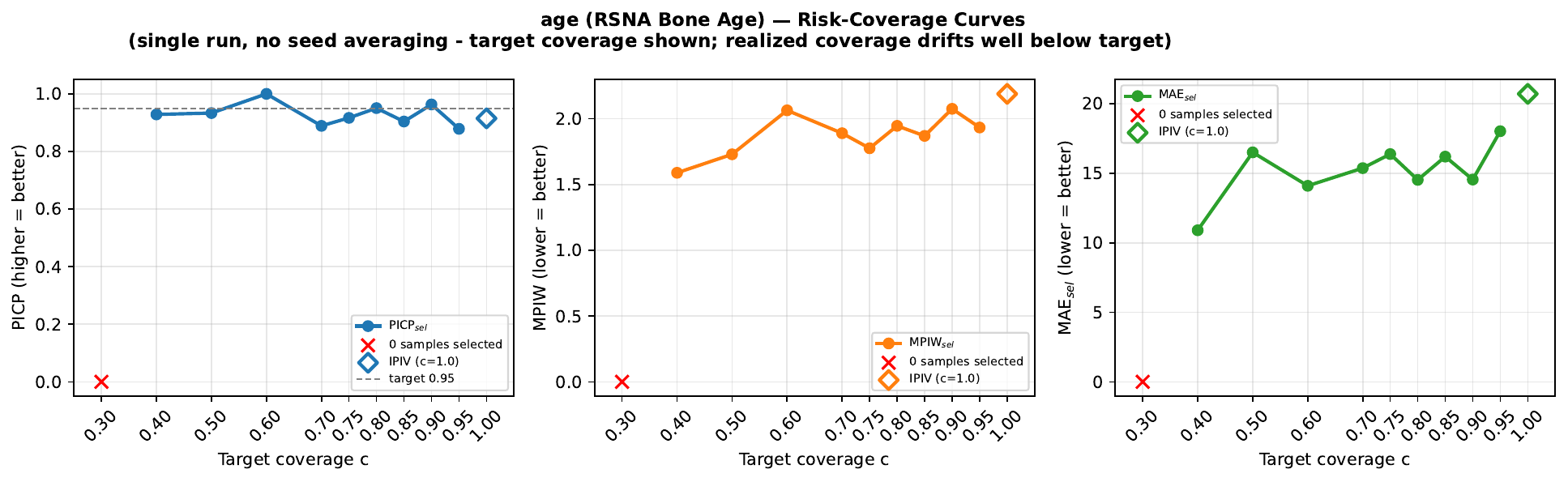}
\includegraphics[width=0.49\textwidth]{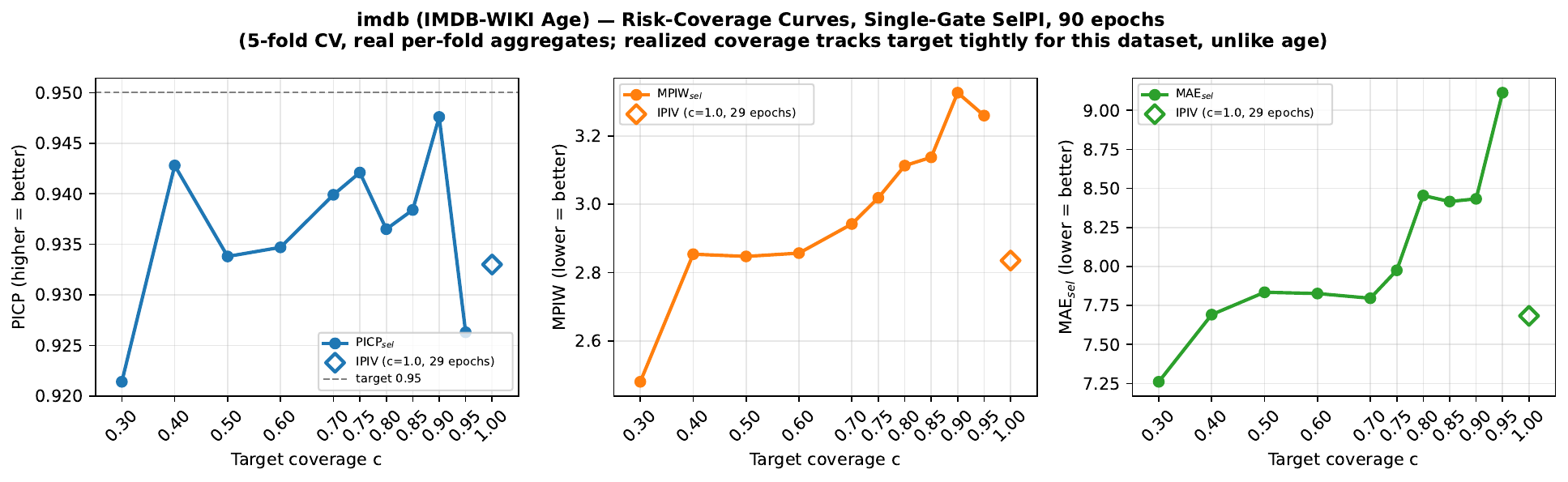}
\end{center}
\caption{\textbf{Risk--coverage curves for the two image tasks}: selective $\picp$, $\mpiw$ and MAE against target acceptance rate, with plain IPIV at full acceptance as the hollow diamond. Note the contrast in calibration transfer: IMDB-WIKI's realized acceptance rate tracks its target almost exactly ($0.70 \to 0.700$, $0.95 \to 0.959$), whereas bone age drifts severely (target $0.90 \to$ realized $0.68$; target $0.30 \to$ realized $0.00$). The cause is sample size --- thresholds calibrated on validation transfer cleanly to IMDB-WIKI's ${\sim}34$k-sample folds and poorly to bone age's 200-image test set. Bone age's ``target coverage'' axis should therefore be read as a training knob, not as a description of what fraction of test inputs was accepted.}
\label{fig:image-rc}
\end{figure}
\FloatBarrier

\begin{table}[h]
\caption{\textbf{Image regression}, selective MAE at $1-\alpha = 0.95$ (bone age in months, IMDB-WIKI in years), grouped by \emph{realized} acceptance rate. Plain IPIV at full acceptance is 20.71 (bone age) and 7.68 (IMDB-WIKI). For the dual gate we give the best and the median trial in each bucket, because its budgets are tuned by search while SelectiveNet is run once at its published settings --- the median is the fair comparison. \emph{Bone age}: every \spp{} variant beats IPIV by a wide margin, and the dual gate's best trial beats SelectiveNet in all three buckets. \emph{IMDB-WIKI}: the dual gate beats IPIV in the two lower buckets, but \textbf{SelectiveNet wins decisively wherever the two overlap}, by 1.3--1.5 years. Rows are keyed by realized rather than target acceptance rate, which matters most for bone age, whose drift is severe (target $0.90$ gives realized $0.68$) because a threshold calibrated on validation transfers poorly to a 200-image test set; IMDB-WIKI has no such problem and is the more trustworthy of the two.}
\label{tab:images}
\begin{center}
\small
\setlength{\tabcolsep}{2.5pt}
\begin{tabular}{l ccc ccc}
\toprule
& \multicolumn{3}{c}{RSNA bone age} & \multicolumn{3}{c}{IMDB-WIKI} \\
\cmidrule(lr){2-4}\cmidrule(lr){5-7}
Realized acceptance rate & $[0.5,0.6)$ & $[0.7,0.8)$ & $[0.8,0.9)$ & $[0.5,0.6)$ & $[0.7,0.8)$ & $[0.8,0.9)$ \\
\midrule
IPIV ($\phi = 1$, no selection) & \multicolumn{3}{c}{20.71} & \multicolumn{3}{c}{7.68} \\
SelectiveNet-regression          & 14.96 & 15.94 & 17.01 & --- & \textbf{6.29} & \textbf{6.76} \\
\sppone{} single gate            & 14.53 & --- & 18.03 & 7.83 & 7.80 & 8.42 \\
\spptwo{} dual gate (median trial)& 15.07 & 16.23 & 17.83 & 7.92 & 8.16 & 8.21 \\
\spptwo{} dual gate (best trial) & \textbf{13.41} & \textbf{15.11} & \textbf{15.50} & \textbf{7.36} & 7.59 & 7.94 \\
\bottomrule
\end{tabular}
\end{center}
\end{table}

\section{Full Per-Dataset Results}
\label{app:full}

\begin{table}[h]
\caption{\textbf{Dual gate vs.\ plain IPIV by acceptance bucket}, median search trial, reduction in selective RMSE; buckets are the realized joint acceptance rate. The gain decays as acceptance rises on every dataset, as it must: as $\phi \to 1$ the accepted set converges on the full test set and there is nothing left to select.}
\label{tab:dualpiven}
\begin{center}
\footnotesize
\begin{tabular}{l rrrr}
\toprule
Dataset & $[0.5,0.6)$ & $[0.6,0.7)$ & $[0.7,0.8)$ & $[0.8,0.9)$ \\
\midrule
yacht & 63.3\% & 61.6\% & 56.1\% & 36.5\% \\
boston & 10.5\% & 8.5\% & 7.5\% & 4.2\% \\
energy & 59.4\% & 51.0\% & 50.5\% & 42.3\% \\
concrete & 33.5\% & 25.5\% & 22.2\% & 16.6\% \\
wine & 16.2\% & 14.8\% & 10.1\% & 8.1\% \\
kin8nm & 11.0\% & 7.1\% & 1.5\% & -6.0\% \\
power-plant & 7.5\% & 4.7\% & 3.2\% & 1.7\% \\
naval & 15.9\% & 14.7\% & 3.0\% & -6.4\% \\
protein & 28.3\% & 22.1\% & 14.4\% & 11.0\% \\
Year-MSD & 43.1\% & 35.3\% & 28.9\% & 21.2\% \\
superconduct & 56.9\% & 47.3\% & 39.5\% & 29.5\% \\
sarcos & 51.5\% & 44.7\% & 35.3\% & 29.4\% \\
diamonds & 73.3\% & 65.0\% & 56.6\% & 44.9\% \\
video\_transcoding & 87.5\% & 84.8\% & 80.1\% & 73.1\% \\
medical\_charges & 30.9\% & 25.7\% & 19.6\% & 16.1\% \\
black\_friday & 8.1\% & 5.4\% & 4.9\% & 4.0\% \\
particulate-matter & 29.7\% & 24.7\% & 19.1\% & 15.1\% \\
nyc-taxi-green & 31.7\% & 28.0\% & 21.8\% & 18.5\% \\
\bottomrule
\end{tabular}
\end{center}
\end{table}
\FloatBarrier

\begin{table}[h]
\caption{\textbf{What abstention buys.} Single-gate \sppone{}, one model trained per acceptance rate (not a searched trial population). Selective RMSE at four acceptance rates, against plain IPIV answering every query; 18 tabular datasets, $1-\alpha = 0.95$. The last column is the reduction at $\phi = 0.5$. Naval's full-acceptance error is $0.001$, so its percentage is a small-denominator artifact rather than a meaningful loss.}
\label{tab:riskcov}
\begin{center}
\footnotesize
\setlength{\tabcolsep}{5pt}
\begin{tabular}{l r rrrr r}
\toprule
& IPIV & \multicolumn{4}{c}{\sppone{} selective RMSE at $\phi =$} & Cut at \\
\cmidrule(lr){3-6}
Dataset & ($\phi=1$) & $0.3$ & $0.5$ & $0.7$ & $0.9$ & $\phi=0.5$ \\
\midrule
yacht & 1.04 & 0.45 & 0.44 & 0.55 & 1.07 & 57.7\% \\
boston & 2.64 & 2.23 & 2.18 & 2.63 & 2.55 & 17.5\% \\
energy & 1.63 & 0.61 & 0.91 & 0.84 & 1.10 & 44.4\% \\
concrete & 6.17 & 3.81 & 4.31 & 5.03 & 5.60 & 30.1\% \\
wine & 0.66 & 0.47 & 0.52 & 0.54 & 0.60 & 21.2\% \\
kin8nm & 0.08 & 0.07 & 0.07 & 0.08 & 0.09 & 13.6\% \\
power-plant & 4.10 & 3.53 & 3.73 & 3.78 & 3.99 & 9.1\% \\
naval & 0.001 & 0.002 & 0.002 & 0.001 & 0.002 & -27.5\% \\
protein & 4.41 & 2.99 & 3.21 & 3.66 & 4.09 & 27.3\% \\
Year-MSD & 9.01 & 4.26 & 5.10 & 6.14 & 7.77 & 43.4\% \\
superconduct & 13.05 & 3.37 & 5.50 & 7.74 & 10.76 & 57.9\% \\
sarcos & 3.23 & 1.10 & 1.42 & 1.90 & 2.59 & 55.9\% \\
diamonds & 563.40 & 103.29 & 137.61 & 221.58 & 350.09 & 75.6\% \\
video\_transcoding & 2.83 & 0.27 & 0.33 & 0.49 & 1.02 & 88.3\% \\
medical\_charges & 0.08 & 0.05 & 0.06 & 0.06 & 0.07 & 30.2\% \\
black\_friday & 0.39 & 0.34 & 0.35 & 0.37 & 0.38 & 8.4\% \\
particulate-matter & 0.39 & 0.22 & 0.27 & 0.31 & 0.35 & 31.2\% \\
nyc-taxi-green & 0.41 & 0.25 & 0.27 & 0.31 & 0.35 & 34.3\% \\
\bottomrule
\end{tabular}
\end{center}
\end{table}
\FloatBarrier

\begin{table}[h]
\caption{\textbf{Dual gate vs.\ SelectiveNet-regression}, selective RMSE at $70\%$ acceptance. ``Best'' is the strongest configuration the search found, ``median'' the typical one; the median column is the fair comparison, because SelectiveNet is run once at its published settings while the dual gate is tuned (\S\ref{sec:setup}). Positive percentages favor \spp{}. Naval and kin8nm are the two clear losses, and both are near-deterministic tasks with little uncertainty for an interval method to characterize.}
\label{tab:vsselnet}
\begin{center}
\footnotesize
\setlength{\tabcolsep}{3.5pt}
\begin{tabular}{l rrr rr}
\toprule
Dataset & SelectiveNet & \spptwo{} best & \spptwo{} med. & Best & Median \\
\midrule
yacht & 1.62 & 0.27 & 0.35 & 83.1\% & 78.3\% \\
boston & 2.42 & 2.20 & 2.39 & 9.2\% & 1.3\% \\
energy & 1.17 & 0.63 & 0.82 & 46.2\% & 30.2\% \\
concrete & 4.66 & 4.24 & 4.59 & 9.1\% & 1.5\% \\
wine & 0.56 & 0.53 & 0.56 & 5.6\% & 0.0\% \\
kin8nm & 0.07 & 0.08 & 0.08 & -9.0\% & -15.5\% \\
power-plant & 3.86 & 3.81 & 3.92 & 1.3\% & -1.5\% \\
naval & 0.0006 & 0.0009 & 0.0011 & -51.3\% & -83.6\% \\
protein & 3.93 & 3.37 & 3.56 & 14.3\% & 9.4\% \\
Year-MSD & 6.13 & 5.71 & 5.96 & 6.8\% & 2.8\% \\
superconduct & 8.63 & 6.81 & 7.40 & 21.1\% & 14.3\% \\
sarcos & 2.40 & 1.76 & 1.91 & 26.4\% & 20.3\% \\
diamonds & 237.56 & 205.32 & 232.45 & 13.6\% & 2.2\% \\
video\_transcoding & 0.79 & 0.43 & 0.50 & 45.3\% & 37.4\% \\
medical\_charges & 0.07 & 0.06 & 0.06 & 12.2\% & 10.2\% \\
black\_friday & 0.38 & 0.35 & 0.37 & 6.2\% & 3.5\% \\
particulate-matter & 0.32 & 0.30 & 0.30 & 6.3\% & 4.1\% \\
nyc-taxi-green & 0.34 & 0.28 & 0.31 & 17.0\% & 9.6\% \\
\bottomrule
\end{tabular}
\end{center}
\end{table}
\FloatBarrier

\begin{figure}[h]
\begin{center}
\includegraphics[width=\textwidth]{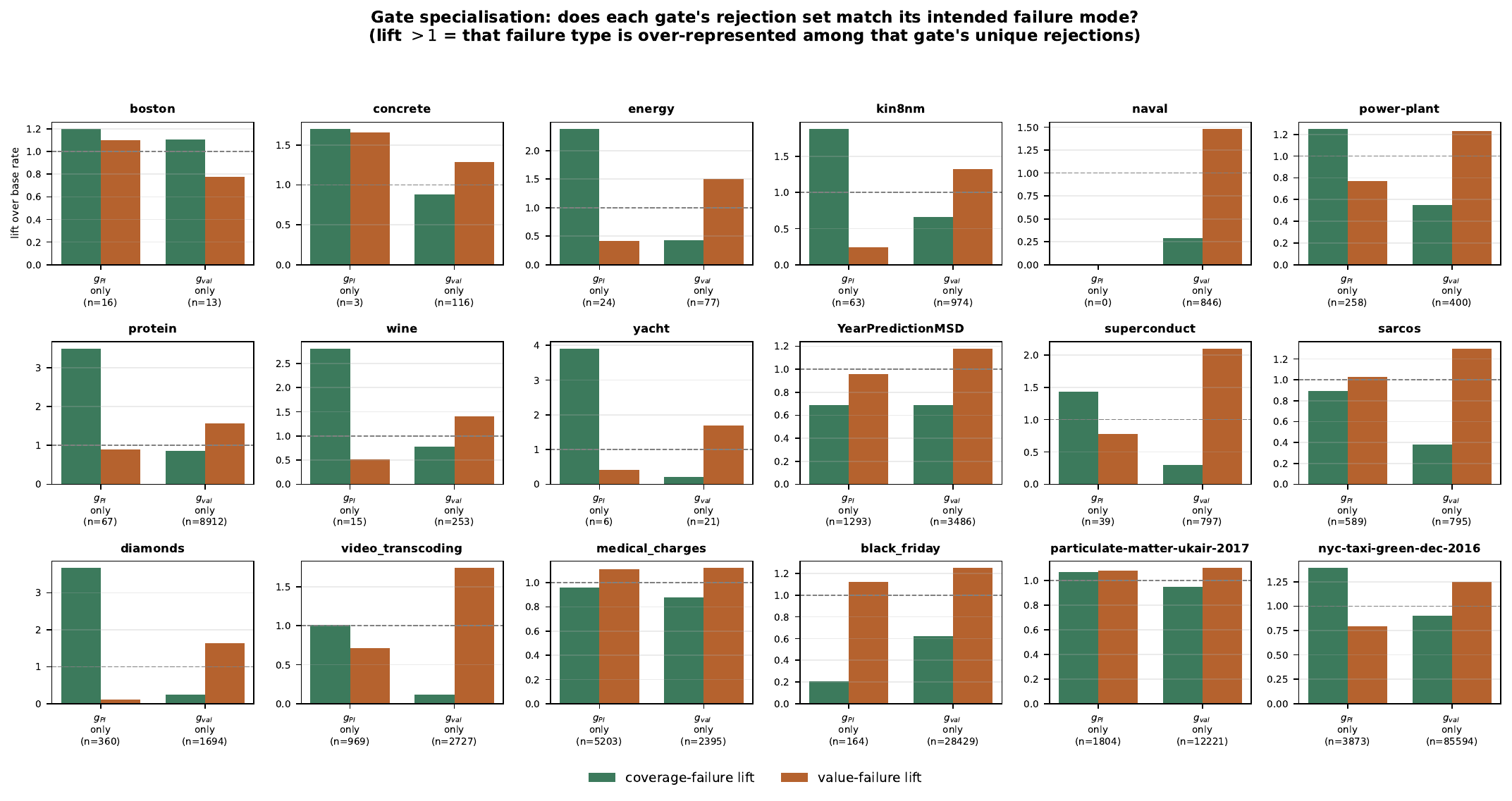}
\end{center}
\caption{\textbf{Gate specialization, all 18 tabular datasets.} For each dataset, the two bars show how over-represented each failure type is among the inputs that \emph{only} $\gpi$ rejects and the inputs that \emph{only} $\gval$ rejects, relative to that dataset's own base rate (dashed line at 1). The signature of successful specialization is a tall green bar on the left pair and a tall orange bar on the right pair. $\gval$ shows it on 17 of 18; $\gpi$ on a clear majority, with the exceptions discussed in \S\ref{sec:complementarity}.}
\label{fig:specialization}
\end{figure}
\FloatBarrier

\begin{table}[h]
\caption{\textbf{Each gate's unique rejections are enriched for the failure mode it was trained on.} For inputs rejected by \emph{only} one gate, how over-represented each failure type is relative to that dataset's base rate; $1.0$ means no enrichment. $\gval$'s unique rejections are enriched for value failure on 17 of 18 datasets; $\gpi$'s are enriched for coverage failure on 13 of the 17 where it has any unique rejections at all (on naval it has none, its rejection set being a subset of $\gval$'s). The value gate is therefore the more consistently load-bearing of the two, which the aggregate results do not reveal.}
\label{tab:speclift}
\begin{center}
\footnotesize
\setlength{\tabcolsep}{5pt}
\begin{tabular}{l rrr rrr}
\toprule
& \multicolumn{3}{c}{rejected by $\gpi$ only} & \multicolumn{3}{c}{rejected by $\gval$ only} \\
\cmidrule(lr){2-4}\cmidrule(lr){5-7}
Dataset & $n$ & cov.\ lift & val.\ lift & $n$ & val.\ lift & cov.\ lift \\
\midrule
yacht & 6 & 3.91 & 0.42 & 21 & 1.69 & 0.22 \\
boston & 16 & 1.20 & 1.10 & 13 & 0.78 & 1.11 \\
energy & 24 & 2.38 & 0.42 & 77 & 1.51 & 0.43 \\
concrete & 3 & 1.70 & 1.66 & 116 & 1.29 & 0.88 \\
wine & 15 & 2.81 & 0.51 & 253 & 1.40 & 0.78 \\
kin8nm & 63 & 1.88 & 0.24 & 974 & 1.32 & 0.66 \\
power-plant & 258 & 1.25 & 0.77 & 400 & 1.23 & 0.55 \\
naval & 0 & --- & --- & 846 & 1.48 & 0.29 \\
protein & 67 & 3.49 & 0.89 & 8912 & 1.56 & 0.85 \\
Year-MSD & 1293 & 0.69 & 0.96 & 3486 & 1.18 & 0.69 \\
superconduct & 39 & 1.43 & 0.78 & 797 & 2.10 & 0.30 \\
sarcos & 589 & 0.89 & 1.03 & 795 & 1.30 & 0.38 \\
diamonds & 360 & 3.67 & 0.12 & 1694 & 1.63 & 0.25 \\
video\_transcoding & 969 & 1.01 & 0.71 & 2727 & 1.74 & 0.12 \\
medical\_charges & 5203 & 0.96 & 1.11 & 2395 & 1.12 & 0.88 \\
black\_friday & 164 & 0.21 & 1.12 & 28429 & 1.25 & 0.62 \\
particulate-matter & 1804 & 1.07 & 1.08 & 12221 & 1.10 & 0.95 \\
nyc-taxi-green & 3873 & 1.39 & 0.79 & 85594 & 1.25 & 0.90 \\
\bottomrule
\end{tabular}
\end{center}
\end{table}
\FloatBarrier

\begin{sidewaystable}
\caption{\textbf{Per-dataset results, merged.} All 18 tabular datasets, $1-\alpha = 0.95$. \emph{$\rho(g,U{-}L)$}: Spearman correlation between the gate and interval width, mean and range over the acceptance grid. \emph{vs.\ IPIV+width}: improvement in selective RMSE over a \emph{separately trained} IPIV with a width threshold --- a cross-pipeline comparison, which is why it goes sharply negative where plain IPIV is simply the better predictor; this is distinct from the same-predictor SE columns, which stay confound-free in $[0,1]$ by construction (\S\ref{sec:setup}). \emph{SE}: selection efficiency of the learned gate over a width threshold on the same base predictor, on selective RMSE and on the interval score (1 = matches the oracle, 0 = matches the width threshold). \emph{$\phi=1$ vs.\ IPIV}: deviation from plain IPIV at full acceptance --- the cost of training with the selective objective. \emph{Ablations}: change in selective RMSE at $\phi=0.5$ when one component is removed or replaced; positive is worse.}
\label{tab:perdataset}
\begin{center}
\footnotesize
\setlength{\tabcolsep}{4pt}
\begin{tabular}{l rl rr cc ccc ccc}
\toprule
& \multicolumn{2}{c}{$\rho(g,\,U-L)$} & \multicolumn{2}{c}{vs.\ IPIV + width} & \multicolumn{2}{c}{Selection efficiency} & \multicolumn{3}{c}{$\phi = 1$ vs.\ IPIV} & \multicolumn{3}{c}{Ablations ($\Delta$RMSE$_{\mathrm{sel}}$)} \\
\cmidrule(lr){2-3}\cmidrule(lr){4-5}\cmidrule(lr){6-7}\cmidrule(lr){8-10}\cmidrule(lr){11-13}
Dataset & mean & range & $\phi=0.5$ & $\phi=0.7$ & RMSE & $\IS_\alpha$ & $\Delta\picp$ & $\Delta\mpiw$ & $\Delta$RMSE & no aux & frozen & $\sqrt{n}$ \\
\midrule
yacht & -0.55 & [-0.76, -0.37] & -14.9\% & 26.7\% & 0.833 & 0.868 & -0.065 & -13.7 & +36.1 & +41.1 & +19.7 & +20.4 \\
boston & -0.18 & [-0.33, -0.04] & 6.3\% & -1.2\% & 0.456 & 0.766 & +0.024 & +20.2 & +4.2 & -4.3 & -7.7 & -3.1 \\
energy & -0.57 & [-0.71, -0.36] & -55.8\% & -12.9\% & 0.732 & 0.951 & +0.005 & +45.6 & +21.4 & -36.8 & +6.8 & +8.1 \\
concrete & -0.13 & [-0.37, 0.25] & 16.0\% & 10.4\% & 0.572 & 0.841 & +0.004 & +5.6 & +4.7 & +1.3 & +6.3 & +1.2 \\
wine & 0.21 & [0.06, 0.48] & 5.3\% & 4.3\% & 0.423 & 0.795 & -0.011 & -9.9 & -3.6 & -1.1 & -0.1 & -2.6 \\
kin8nm & -0.14 & [-0.50, 0.35] & -13.3\% & -12.6\% & 0.710 & 0.884 & -0.014 & +38.3 & +51.7 & +1.8 & +11.1 & -0.8 \\
power-plant & -0.23 & [-0.47, 0.24] & 3.0\% & 0.6\% & 0.160 & 0.526 & -0.009 & -2.2 & +0.1 & +2.8 & +2.9 & +0.6 \\
naval & -0.37 & [-0.70, 0.06] & -255.0\% & -110.7\% & 0.891 & 0.972 & -0.063 & +91.0 & +103.1 & +26.7 & -20.0 & -6.7 \\
protein & -0.15 & [-0.39, 0.40] & 12.0\% & 8.0\% & 0.602 & 0.866 & -0.025 & -6.7 & +2.0 & +4.1 & +4.4 & +1.4 \\
Year-MSD & -0.56 & [-0.73, -0.01] & 10.0\% & 12.3\% & 0.567 & 0.799 & -0.026 & -11.9 & -0.7 & +1.9 & +3.9 & -0.2 \\
superconduct & -0.32 & [-0.67, 0.54] & 9.3\% & 14.9\% & 0.868 & 0.981 & -0.031 & -16.5 & -0.1 & +41.2 & +13.2 & -2.7 \\
sarcos & -0.63 & [-0.74, -0.38] & 23.1\% & 18.2\% & 0.543 & 0.876 & -0.011 & -7.1 & +0.7 & +2.4 & +33.3 & -0.0 \\
diamonds & -0.37 & [-0.85, 0.44] & -1.5\% & 4.1\% & 0.949 & 0.997 & -0.012 & -8.5 & +11.0 & +33.3 & +8.2 & -12.2 \\
video\_transcoding & -0.27 & [-0.70, 0.58] & 53.3\% & 40.1\% & 0.987 & 0.998 & -0.019 & -14.2 & +35.9 & +31.2 & +54.2 & +39.5 \\
medical\_charges & -0.71 & [-0.88, -0.38] & 3.7\% & 1.8\% & 0.321 & 0.661 & -0.006 & -3.3 & +0.4 & +0.4 & +4.9 & +0.4 \\
black\_friday & 0.19 & [-0.32, 0.53] & 8.2\% & 4.1\% & 0.694 & 0.959 & -0.054 & -18.5 & +28.9 & +0.6 & +3.7 & -0.4 \\
particulate-matter & -0.77 & [-0.85, -0.63] & 5.7\% & 3.6\% & 0.147 & 0.337 & -0.004 & -3.6 & -0.6 & -0.1 & +3.7 & +0.7 \\
nyc-taxi-green & -0.05 & [-0.26, 0.11] & 34.0\% & 22.0\% & 0.785 & 0.959 & -0.028 & -10.5 & +10.5 & +2.1 & +12.9 & +2.0 \\
\bottomrule
\end{tabular}
\end{center}
\end{sidewaystable}

\section{Statistical Tests}
\label{app:stats}
Two comparisons are kept separate because only the second isolates the selector. The learned gate against a width threshold \emph{on the same predictor}: 20 wins, 0 losses, $p = 2\times10^{-6}$, median selection efficiency $0.587$. The whole pipeline against a separately trained IPIV + width: 16 wins, 3 losses, $p = 5.2\times10^{-4}$. On the interval score the same-predictor comparison gives 20 wins, 0 losses, median $0.872$. The gate against deep-ensemble disagreement (a cross-pipeline comparison of selection efficiency): 20 wins, 0 losses ($p = 2\times10^{-6}$) on selective RMSE and 19 wins, 0 losses ($p = 4\times10^{-6}$) on the interval score, for both disagreement signals. Coverage within width quintiles of the accepted set~\citep{angelopoulos2023conformal} is nearly uniform on the six largest datasets (gap $\le 0.017$) but not on the small UCI sets (up to $0.375$ on yacht), where accepted-set calibration is partly carried by one easy sub-population. Friedman and Nemenyi tests are not applicable, since they rank a field of methods and each of our comparisons has two.

\section{Two-Budget Frontier}
\label{app:frontier}

\begin{figure}[h]
\begin{center}
\includegraphics[width=0.62\textwidth]{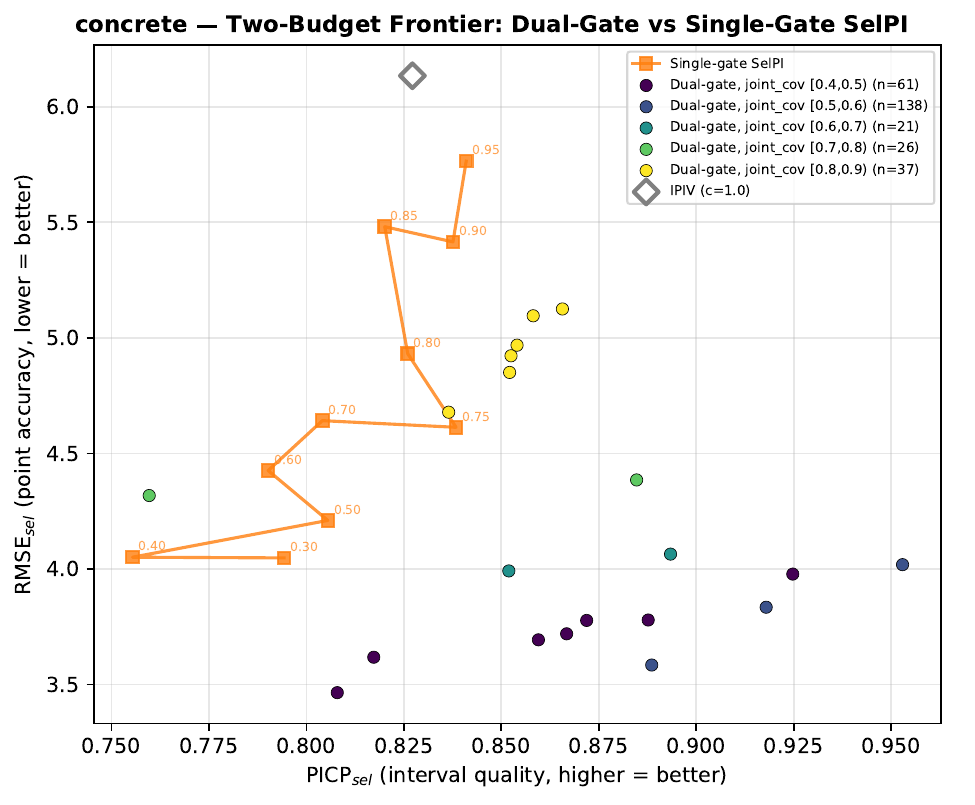}
\end{center}
\caption{\textbf{The two-budget frontier}, Concrete. Each dot is one dual-gate configuration, colored by the joint acceptance rate it actually realized; the orange line is the single-gate model traced over its acceptance grid. Down and to the right is better (lower selective error, higher selective interval coverage). The dual-gate cloud sits below and to the right of the single-gate curve at matched acceptance --- the trade-off a single scalar gate cannot express. Per-dataset versions for all 18 tabular datasets are in Figure~\ref{fig:grid-frontier}.}
\label{fig:frontier}
\end{figure}
\section{``When Does It Help'': Regression Details}
\label{app:when}
\begin{figure}[h]
\begin{center}
\includegraphics[width=\textwidth]{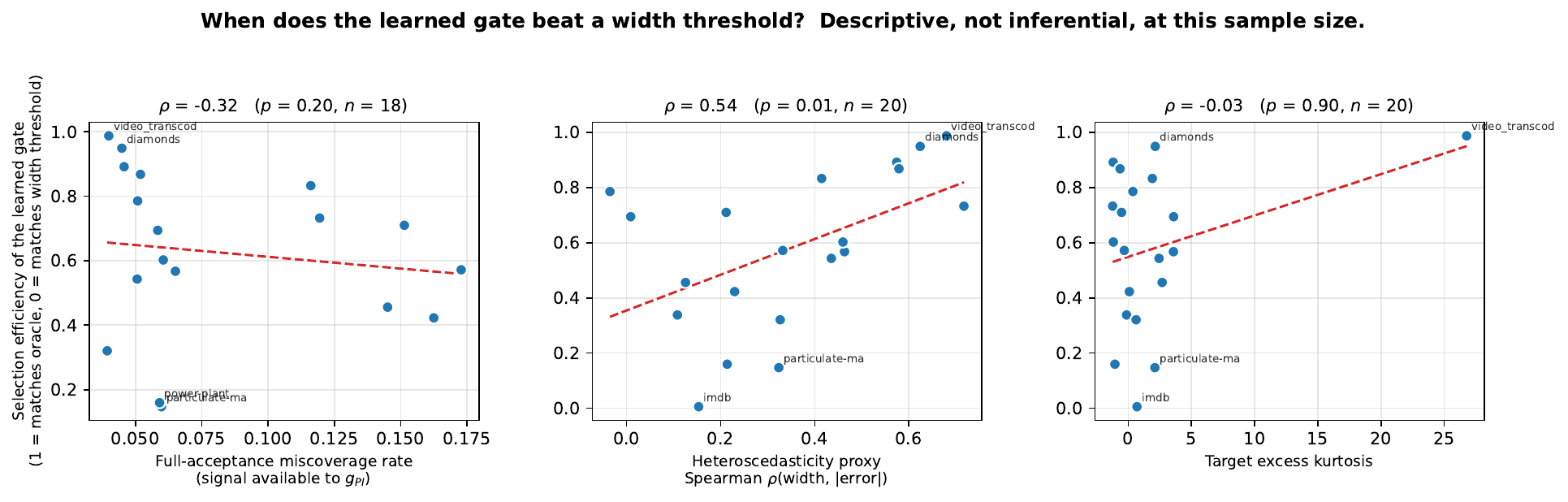}
\end{center}
\caption{\textbf{When joint selection helps.} Per-dataset selection efficiency of the learned gate against (left) the full-acceptance miscoverage rate, (middle) a heteroscedasticity statistic, (right) target excess kurtosis. Each point is one of 20 datasets (18 for the left panel, where the two image tasks lack a full-acceptance miscoverage figure); Spearman $\rho$ and its $p$-value are in each panel title, with a least-squares line for orientation only. Only the heteroscedasticity relationship is distinguishable from noise at this sample size.}
\label{fig:when}
\end{figure}
The figure supports only one of the three relationships \S\ref{sec:when} predicts. \textbf{Heteroscedasticity: confirmed}, $\rho = 0.54$, $p = 0.013$, $n = 20$, where the heteroscedasticity statistic is Spearman $\rho(U-L, |y - \yhat|)$ on the plain IPIV baseline. \textbf{Excess kurtosis: no relationship}, $\rho = -0.03$, $p = 0.91$. \textbf{Full-acceptance miscoverage: weakly negative and not significant}, $\rho = -0.32$, $p = 0.20$ --- if anything the gate does \emph{worse} where there is more miscoverage, the opposite of the intuitive direction.

\section{Distribution-Shift Protocols and Results}
\label{app:shift}
For each dataset and run we regenerate the exact test split the model saw, identify the covariate that best tracks realized error, and treat its top quintile as a shifted tail; no retraining is needed. Holding each selector's threshold fixed at its $\phi = 0.7$ value, we measure how much its abstention rate rises on the tail. The learned gate raises abstention by a median $+0.193$ against $+0.046$ for the width threshold, and reacts more strongly on 15 of 18 datasets; on diamonds, black\_friday and nyc-taxi-green the gate abstains on $73$--$96\%$ of the tail while the width threshold's rate moves by less than $0.3$, sometimes in the \emph{wrong} direction. On two datasets (naval, particulate-matter) both selectors abstain \emph{less} on the tail, i.e.\ the chosen covariate isolates an easier region there. Because the tail is constructed from a held-out covariate rather than injected, the protocol does not let us claim that the advantage grows with shift severity, and it does not cover the image tasks.

\section{Full Ablation Results}
\label{app:ablations}
\begin{table}[h]
\caption{\textbf{Ablations}: change in selective RMSE at $\phi = 0.5$ vs.\ the full recipe (positive: ablation is worse); 18 tabular datasets. W/B: datasets worse/better.}
\label{tab:ablations}
\centering
\footnotesize
\setlength{\tabcolsep}{3pt}
\begin{tabular}{@{}l r c r@{}}
\toprule
Variant & Med. & W/B & Range (\%) \\
\midrule
no auxiliary heads          & $+2.0\%$ & 14/4 & $-37$, $+41$ \\
frozen IPIV, post-hoc gate  & $+5.6\%$ & 15/3 & $-20$, $+54$ \\
$\sqrt{n}$ for $\sqrt{S_g}$ & $+0.2\%$ & 9/9  & $-12$, $+39$ \\
\bottomrule
\end{tabular}
\end{table}

\begin{figure}[h]
\begin{center}
\includegraphics[width=\textwidth]{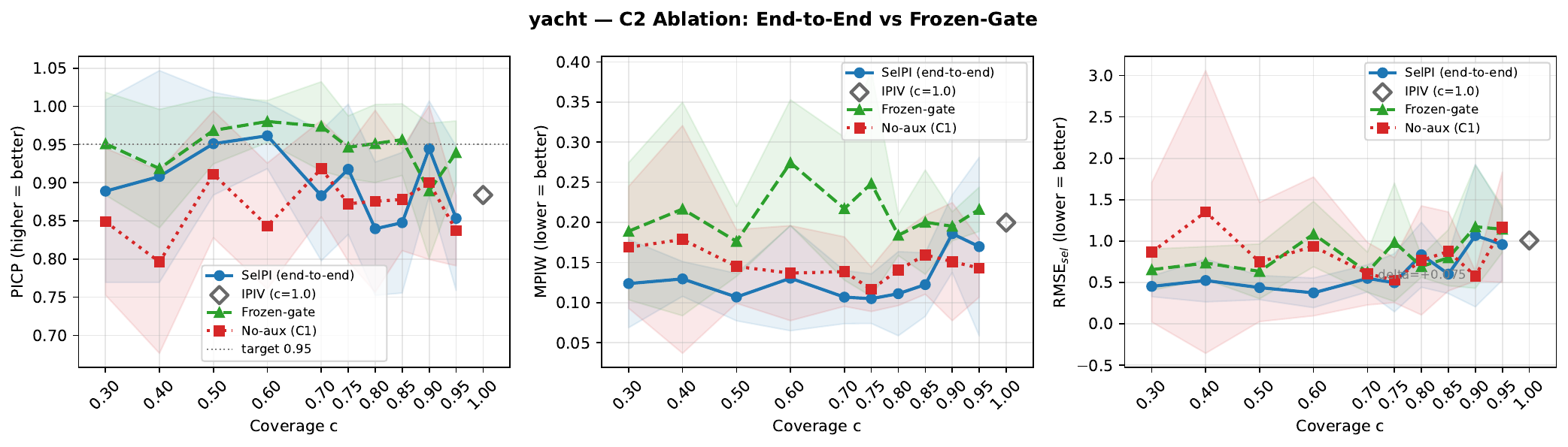}
\end{center}
\caption{\textbf{Ablations on Yacht}, the dataset where the effects are largest: end-to-end \spp{} against a gate trained on a frozen IPIV model, and against removing the auxiliary head. Selective $\picp$, $\mpiw$ and RMSE against target acceptance rate; shaded bands are $\pm 1$ s.d.\ over 5 runs. End-to-end training dominates on both interval width and selective error across the whole grid. Equivalent figures exist for all 18 tabular datasets.}
\label{fig:ablation}
\end{figure}
\begin{figure}[h]
\begin{center}\includegraphics[width=0.86\textwidth]{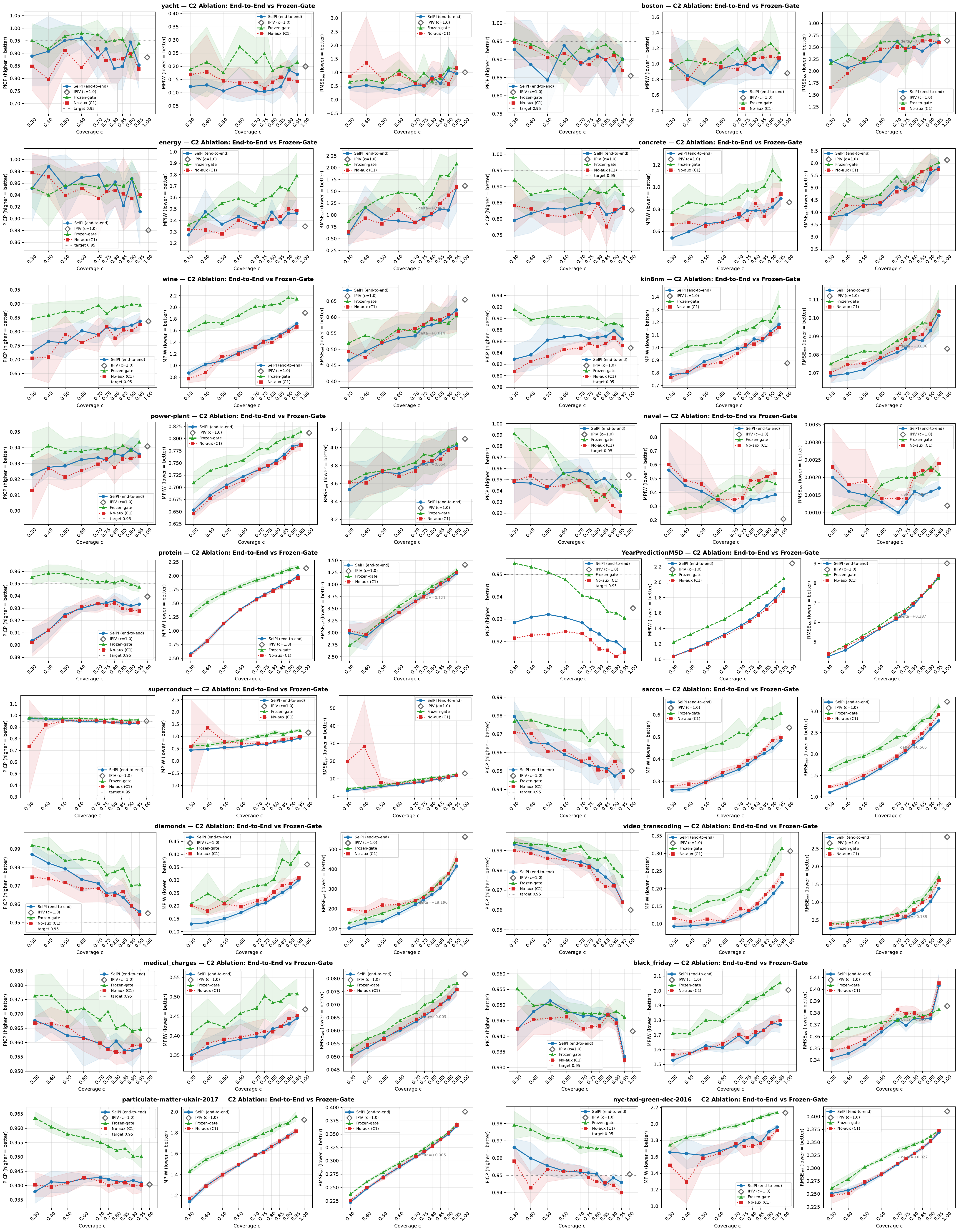}\end{center}
\caption{\textbf{Ablation: end-to-end vs.\ frozen-gate training}, with the no-auxiliary-head variant overlaid. Frozen-gate is worse on 15 of 18 datasets (Table~\ref{tab:ablations}). This overlay also contains the no-auxiliary-head comparison shown on its own in Figure~\ref{fig:grid-c1} below.}
\label{fig:grid-c2}
\end{figure}
\begin{figure}[h]
\begin{center}\includegraphics[width=0.86\textwidth]{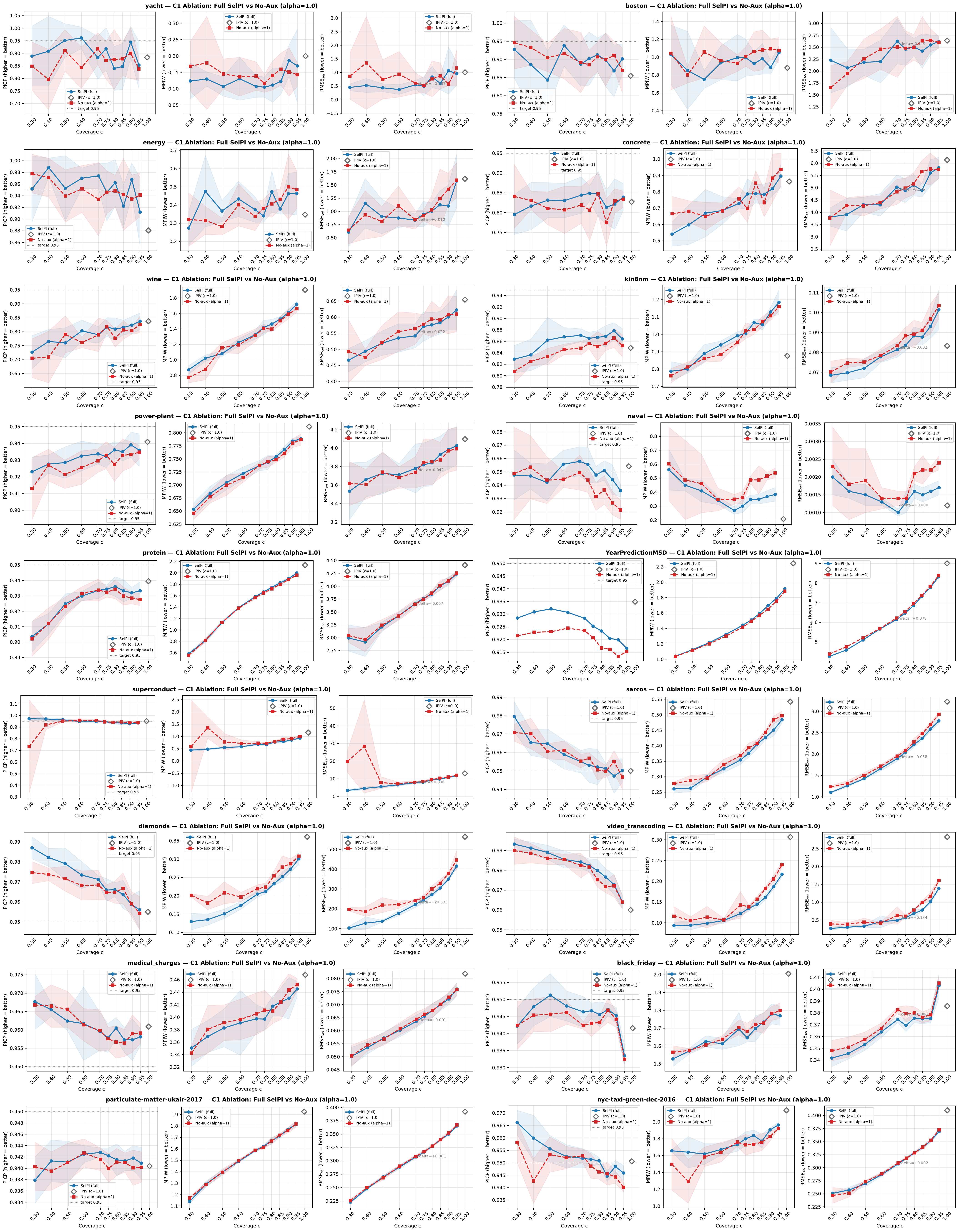}\end{center}
\caption{\textbf{Ablation: with vs.\ without the auxiliary head}, isolated into its own panel (also visible as part of the overlay in Figure~\ref{fig:grid-c2}).}
\label{fig:grid-c1}
\end{figure}
\begin{figure}[h]
\begin{center}\includegraphics[width=0.86\textwidth]{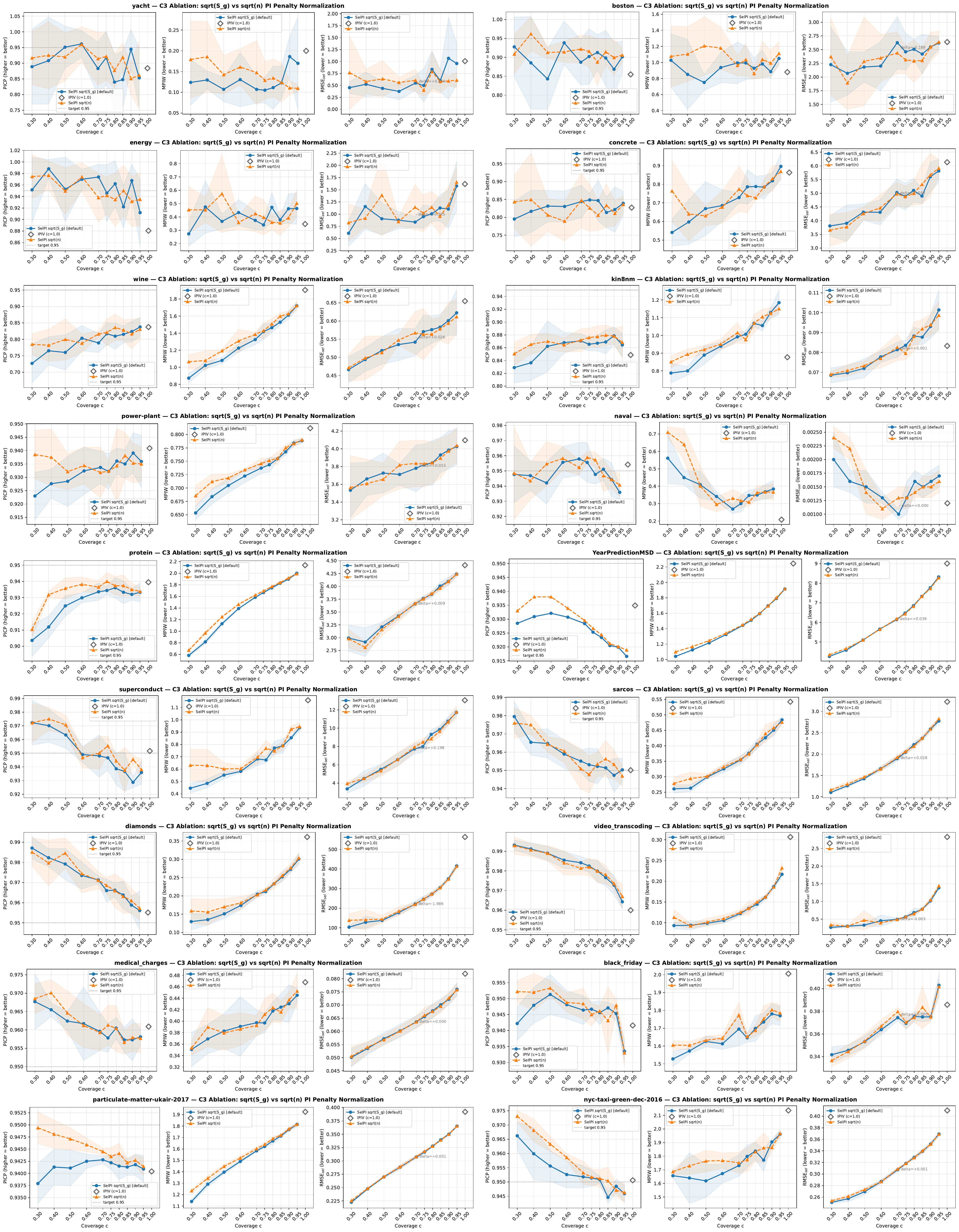}\end{center}
\caption{\textbf{Ablation: $\sqrt{S_g}$ vs.\ $\sqrt{n}$ penalty normalization.} The curves track each other closely on most datasets, which is the visual form of the null result in Table~\ref{tab:ablations}.}
\label{fig:grid-c3}
\end{figure}

The three ablations of Table~\ref{tab:ablations} were run on all 18 tabular datasets; per-dataset values are in Table~\ref{tab:perdataset} and Figures~\ref{fig:grid-c2}--\ref{fig:grid-c3}.

\section{Per-Dataset Figures}
\label{app:gallery}
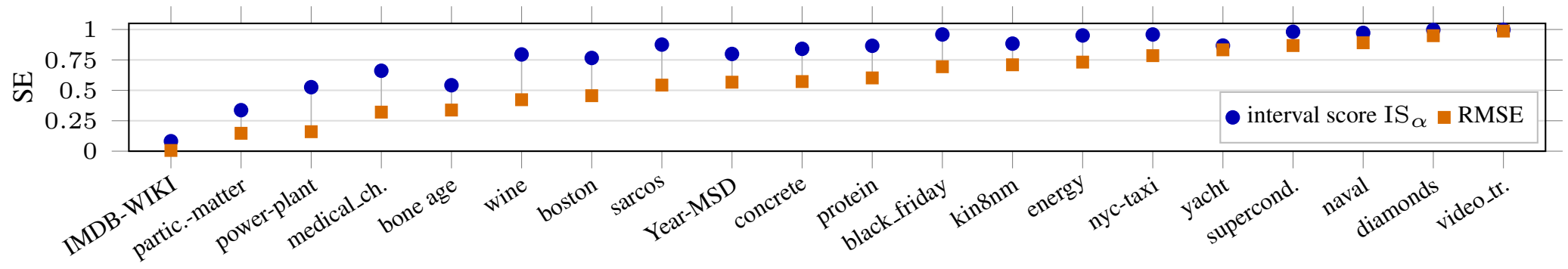
\begin{figure}[h]
\centering
\begin{tikzpicture}
\begin{axis}[
  width=\textwidth, height=2.7cm,
  ymin=0, ymax=1.05, ytick={0,0.25,0.5,0.75,1},
  ylabel={SE}, ylabel style={font=\scriptsize, yshift=-4pt},
  ymajorgrids, grid style={gray!25},
  xmin=0.4, xmax=20.6, xtick={1,...,20},
  xticklabels={IMDB-WIKI,partic.-matter,power-plant,medical\_ch.,bone age,wine,boston,sarcos,Year-MSD,concrete,protein,black\_friday,kin8nm,energy,nyc-taxi,yacht,supercond.,naval,diamonds,video\_tr.},
  x tick label style={font=\tiny, rotate=32, anchor=north east, inner sep=1pt},
  y tick label style={font=\tiny},
  legend style={font=\tiny, at={(0.995,0.06)}, anchor=south east, legend columns=2, draw=gray!50, inner sep=1.5pt},
  tick align=outside, tick style={gray},
]
\addplot[only marks, mark=*, mark size=1.6pt, blue!70!black] coordinates {
(1,0.082)(2,0.337)(3,0.526)(4,0.661)(5,0.542)(6,0.795)(7,0.766)(8,0.876)(9,0.799)(10,0.841)(11,0.866)(12,0.959)(13,0.884)(14,0.951)(15,0.959)(16,0.868)(17,0.981)(18,0.972)(19,0.997)(20,0.998)};
\addlegendentry{interval score $\IS_\alpha$}
\addplot[only marks, mark=square*, mark size=1.4pt, orange!85!black] coordinates {
(1,0.006)(2,0.147)(3,0.160)(4,0.321)(5,0.338)(6,0.423)(7,0.456)(8,0.543)(9,0.567)(10,0.572)(11,0.602)(12,0.694)(13,0.710)(14,0.732)(15,0.785)(16,0.833)(17,0.868)(18,0.891)(19,0.949)(20,0.987)};
\addlegendentry{RMSE}
\foreach \x/\a/\b in {1/0.006/0.082,2/0.147/0.337,3/0.160/0.526,4/0.321/0.661,5/0.338/0.542,6/0.423/0.795,7/0.456/0.766,8/0.543/0.876,9/0.567/0.799,10/0.572/0.841,11/0.602/0.866,12/0.694/0.959,13/0.710/0.884,14/0.732/0.951,15/0.785/0.959,16/0.833/0.868,17/0.868/0.981,18/0.891/0.972,19/0.949/0.997,20/0.987/0.998}{
  \edef\temp{\noexpand\draw[gray!60, line width=0.3pt] (axis cs:\x,\a) -- (axis cs:\x,\b);}\temp}
\end{axis}
\end{tikzpicture}
\caption{\textbf{Selection efficiency per dataset} (single-gate \sppone{}): the share of the gap between the same-predictor width threshold ($\mathrm{SE}=0$) and the oracle ($\mathrm{SE}=1$) that the learned gate closes, on selective RMSE and on the selective interval score. The gate beats the width threshold on every dataset on both metrics, and by more on the interval score on all 20. The two image tasks are illustrative (\S\ref{sec:setup}); SE is scale-free, so their real-unit widths do not affect comparability.}
\label{fig:se}
\end{figure}

Each grid shows the same plot for all 18 tabular datasets, in the order used throughout the appendix: the nine UCI sets, then the nine larger ones.

\begin{figure}[h]
\begin{center}\includegraphics[width=0.86\textwidth]{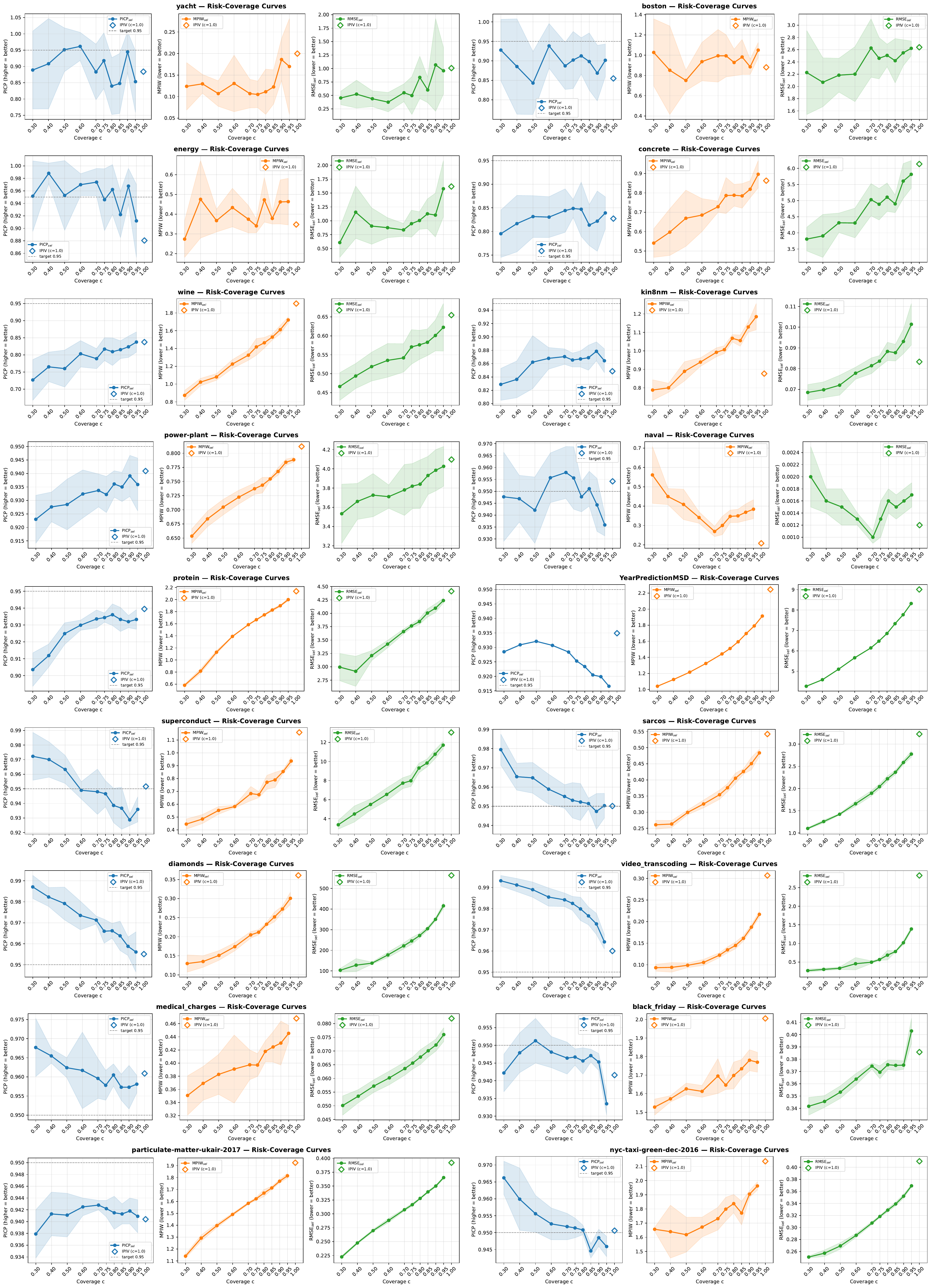}\end{center}
\caption{\textbf{Risk--coverage curves}: selective $\picp$, $\mpiw$ and RMSE against target acceptance rate, with plain IPIV at full acceptance as the hollow diamond. Bands are $\pm 1$ s.d.\ over runs.}
\label{fig:grid-rc}
\end{figure}
\FloatBarrier

\begin{figure}[h]
\begin{center}\includegraphics[width=0.95\textwidth]{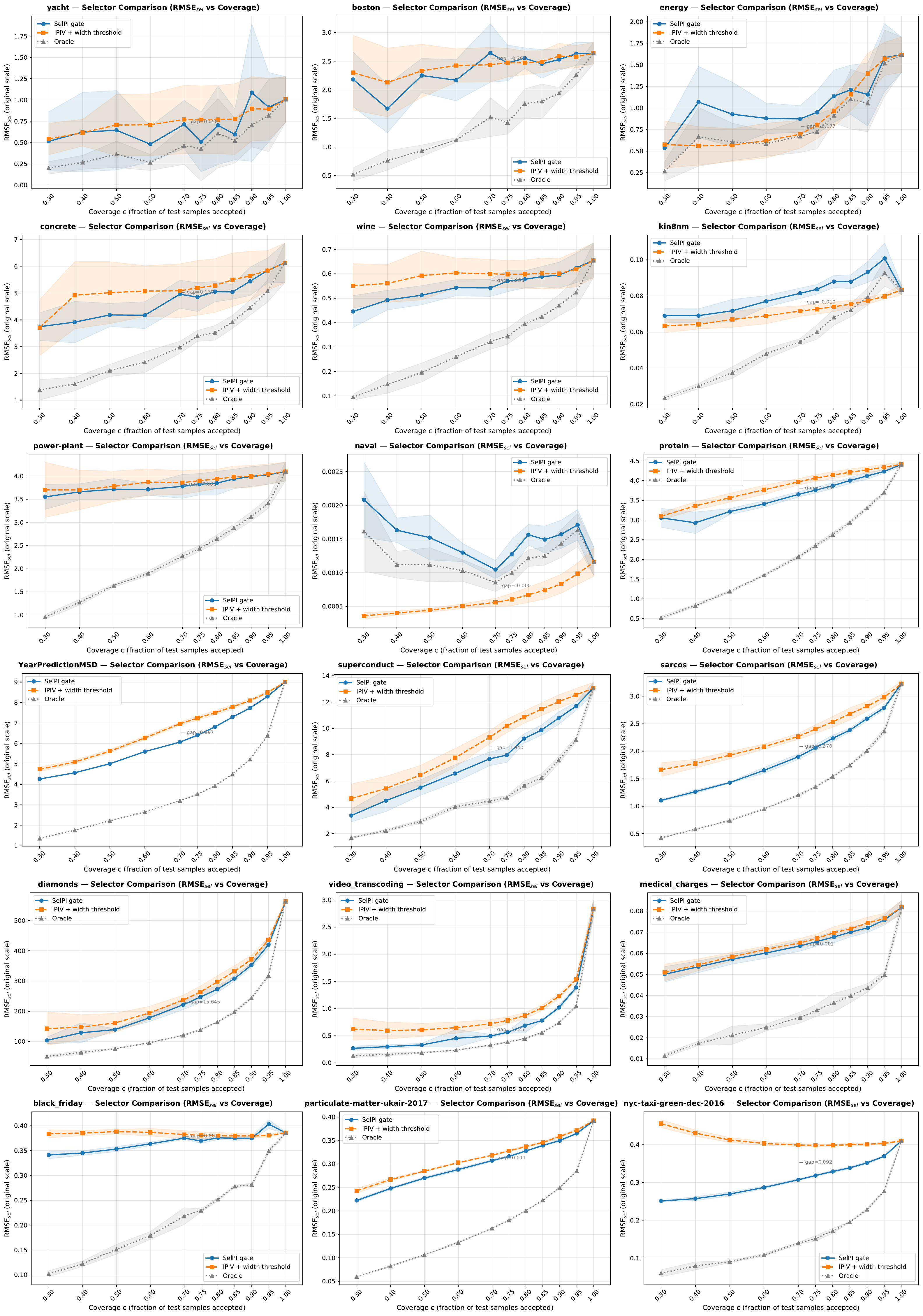}\end{center}
\caption{\textbf{Selector comparison}: the learned gate against IPIV + width threshold and against an oracle ranking by realized error. The persistent oracle gap is the headroom that remains on every dataset.}
\label{fig:grid-sel}
\end{figure}
\FloatBarrier

\begin{figure}[h]
\begin{center}\includegraphics[width=0.86\textwidth]{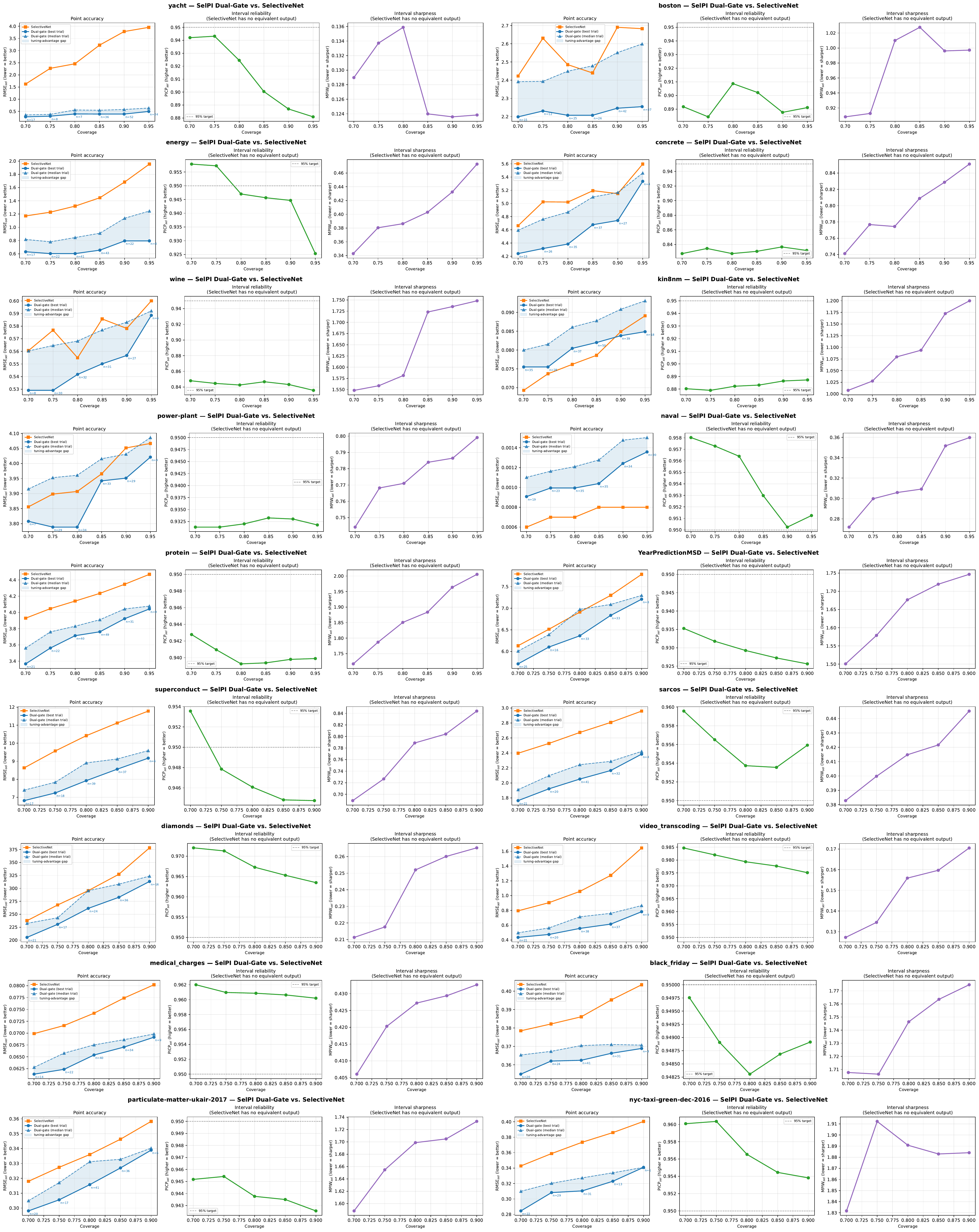}\end{center}
\caption{\textbf{Dual gate vs.\ SelectiveNet-regression}: point accuracy, then the interval reliability and sharpness SelectiveNet cannot produce at all. Shading between best and median trial is the tuning advantage of \S\ref{sec:setup}.}
\label{fig:grid-selnet}
\end{figure}
\FloatBarrier

\begin{figure}[h]
\begin{center}\includegraphics[width=0.95\textwidth]{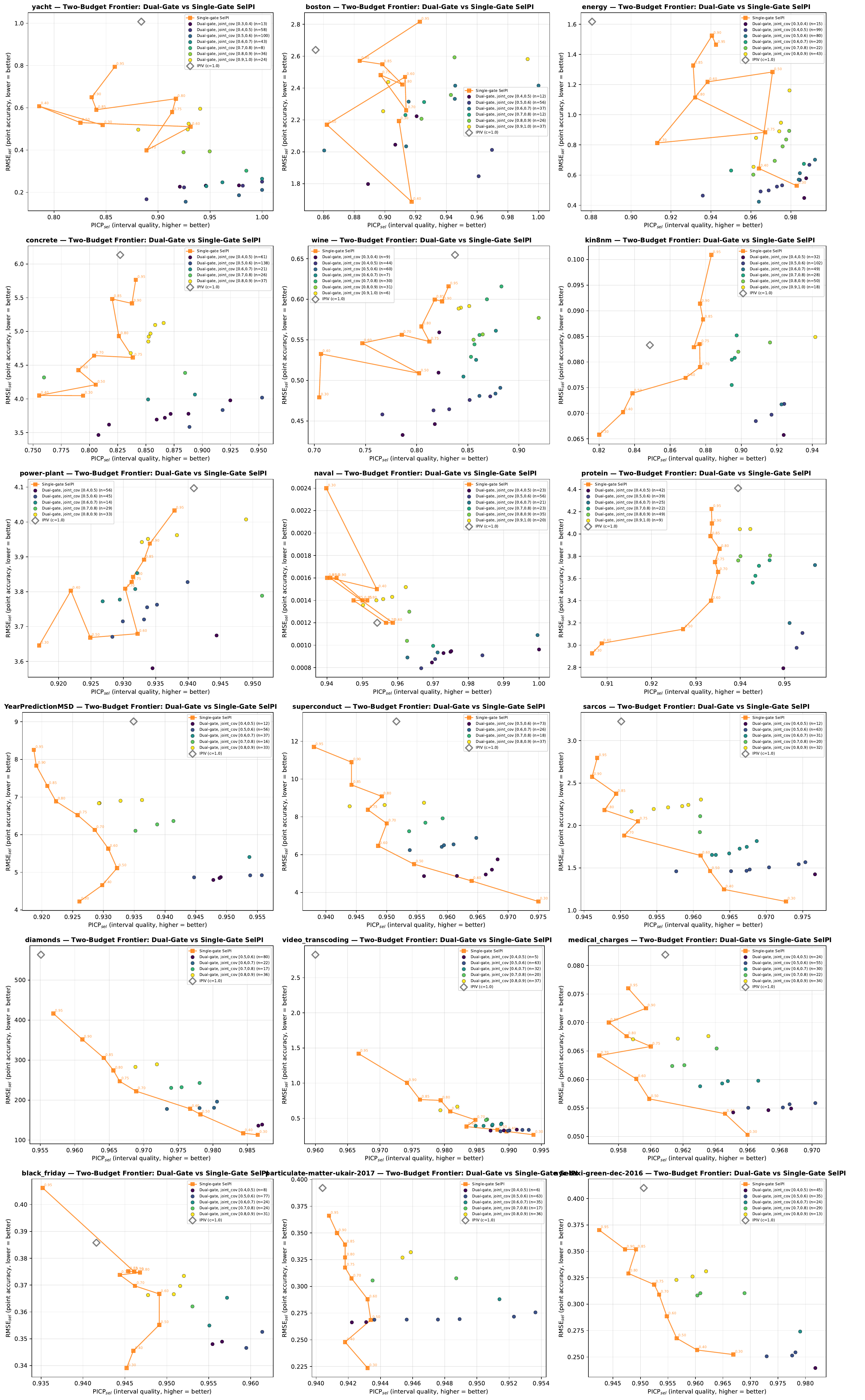}\end{center}
\caption{\textbf{Two-budget frontier}: every dual-gate configuration, colored by realized joint acceptance rate, against the single-gate curve. Down and to the right is better.}
\label{fig:grid-frontier}
\end{figure}
\FloatBarrier

\FloatBarrier

\FloatBarrier

\section{Full-Acceptance Baselines from the Literature}
\label{app:published}
Our evaluation compares selectors, so it needs baselines that abstain. Neither parent paper does: IPIV and its comparators answer every query. Their published numbers are nonetheless useful in two ways, and we reproduce them here rather than leave the impression that no wider comparison exists. First, they place our reproduction of IPIV --- the base predictor every \spp{} number is built on --- against the numbers its authors report, on the same datasets and protocol. Second, they show where the interval-quality field stood at full acceptance before a reject option is added.

\begin{table}[h]
\caption{Full-acceptance ($\phi = 1$) results as published in the IPIV paper, with our own reproduction of IPIV alongside (the \emph{repro} columns --- this is plain IPIV retrained in our environment, not \spp{}). QD, QD+ and DE are that paper's own baselines. \emph{Ens.} is our own $M{=}5$-model IPIV ensemble (\S\ref{sec:setup}), the base predictor behind the deep-ensemble selector of \S\ref{sec:results}. Published figures use 20 splits and 5-model ensembles; ours use 5 splits (and, for \emph{Ens.}, 5 members per split), so small differences are expected and the comparison is one of fidelity, not of ranking. RMSE/MAE are on the original target scale; MPIW is in standardized units.}
\label{tab:published}
\begin{center}
\scriptsize
\setlength{\tabcolsep}{2.6pt}
\begin{tabular}{l cccccc c cccccc}
\toprule
& \multicolumn{6}{c}{RMSE} & & \multicolumn{6}{c}{MPIW} \\
\cmidrule(lr){2-7}\cmidrule(lr){9-14}
Dataset & DE & QD & QD+ & IPIV & \emph{repro} & \emph{Ens.} & & DE & QD & QD+ & IPIV & \emph{repro} & \emph{Ens.} \\
\midrule
Boston      & 2.87 & 3.39 & 6.12  & 3.13 & \emph{2.64}  & \emph{2.70} & & 0.87 & 1.15 & 4.98 & 1.09 & \emph{0.88} & \emph{1.09} \\
Concrete    & 5.21 & 5.88 & 11.87 & 5.43 & \emph{6.17}  & \emph{5.79} & & 1.01 & 1.08 & 3.43 & 1.02 & \emph{0.86} & \emph{1.08} \\
Energy      & 1.68 & 2.28 & 7.05  & 1.65 & \emph{1.63}  & \emph{1.70} & & 0.49 & 0.45 & 1.52 & 0.42 & \emph{0.35} & \emph{0.42} \\
Kin8nm      & 0.08 & 0.08 & 0.19  & 0.07 & \emph{0.08}  & \emph{0.08} & & 1.14 & 1.18 & 2.64 & 1.10 & \emph{0.88} & \emph{1.10} \\
Naval       & 0.00 & 0.00 & 0.01  & 0.00 & \emph{0.00}  & \emph{0.00} & & 0.31 & 0.27 & 2.85 & 0.24 & \emph{0.21} & \emph{0.27} \\
Power       & 3.99 & 4.14 & 12.00 & 4.08 & \emph{4.10}  & \emph{4.09} & & 0.91 & 0.86 & 1.51 & 0.86 & \emph{0.81} & \emph{0.86} \\
Protein     & 4.36 & 4.99 & 5.10  & 4.35 & \emph{4.41}  & \emph{4.36} & & 2.68 & 2.27 & 3.30 & 2.26 & \emph{2.14} & \emph{2.25} \\
Wine        & 0.62 & 0.67 & 0.70  & 0.63 & \emph{0.66}  & \emph{0.62} & & 2.50 & 2.24 & 4.89 & 2.22 & \emph{1.91} & \emph{2.20} \\
Yacht       & 1.38 & 1.10 & 8.19  & 0.98 & \emph{1.04}  & \emph{0.95} & & 0.33 & 0.18 & 1.25 & 0.17 & \emph{0.20} & \emph{0.28} \\
Year-MSD    & 8.95 & 9.30 & 10.11 & 8.93 & \emph{9.01}  & \emph{8.93} & & 2.91 & 2.45 & 4.25 & 2.42 & \emph{2.25} & \emph{2.47} \\
\midrule
IMDB-WIKI (MAE) & 6.66 & 10.23 & 10.08 & 7.03 & \emph{7.68} & \emph{7.42} & & 2.61 & 3.47 & 5.01 & 2.87 & \emph{2.84} & \emph{3.28} \\
Bone age (MAE)  & 18.69 & 20.24 & 25.18 & 18.13 & \emph{20.71} & \emph{19.46} & & 2.17 & 1.99 & 5.24 & 2.09 & \emph{2.19} & \emph{3.38} \\
\bottomrule
\end{tabular}
\end{center}
\end{table}

Our IPIV reproduction tracks the published numbers closely on the tabular sets and is somewhat worse on the two image tasks, where we train a single model rather than an ensemble. Our own \emph{Ens.} column closes most of that gap: on IMDB-WIKI it improves \emph{repro}'s MAE from $7.68$ to $7.42$, and on bone age from $20.71$ to $19.46$, tracking the published DE/IPIV figures more closely once the base predictor is itself an ensemble rather than a single model. It still trails the published DE figure on IMDB-WIKI ($6.66$), which is expected: DE is a different architecture (a Gaussian-NLL ensemble) evaluated over 20 splits, ours is an IPIV ensemble over 5. On the tabular sets the \emph{Ens.} MPIW column tracks the published IPIV column almost exactly, as it should, since the published IPIV figures are themselves five-model ensembles. This \emph{Ens.} predictor is also the one behind the deep-ensemble selector baseline of \S\ref{sec:results}--\S\ref{sec:analysis}, which we compare directly against \spp{}'s learned gates.

\section{Deep-Ensemble Selector: Per-Dataset Results}
\label{app:ensemble}
Table~\ref{tab:ensemble} gives the per-dataset values behind the selection-efficiency columns of Table~\ref{tab:auroc}: the selection efficiency of the learned gate and of the two disagreement selectors on both metrics, each against its own predictor's width threshold and oracle, and the two failure-mode AUROCs of the disagreement signals.
\begin{table}[h]
\caption{Per-dataset selection efficiency (SE) on selective RMSE and on the interval score for the learned gate (\sppone{}) and for the two deep-ensemble disagreement selectors, and AUROC of each disagreement signal for coverage failure (cov.) and value failure (val.). Each SE is computed against its own predictor's width threshold and oracle; ``--'' marks an undefined SE (width and oracle AURC coincide).}
\label{tab:ensemble}
\begin{center}
\scriptsize
\setlength{\tabcolsep}{3pt}
\resizebox{\textwidth}{!}{%
\begin{tabular}{l ccc ccc cc cc}
\toprule
& \multicolumn{3}{c}{SE, selective RMSE} & \multicolumn{3}{c}{SE, interval score} & \multicolumn{2}{c}{AUROC, point disagr.} & \multicolumn{2}{c}{AUROC, bound disagr.} \\
\cmidrule(lr){2-4}\cmidrule(lr){5-7}\cmidrule(lr){8-9}\cmidrule(lr){10-11}
Dataset & gate & point & bound & gate & point & bound & cov. & val. & cov. & val. \\
\midrule
yacht & 0.833 & $-0.961$ & $-1.062$ & 0.868 & -- & -- & 0.761 & 0.735 & 0.840 & 0.747 \\
boston & 0.456 & $0.108$ & $-0.001$ & 0.766 & $0.225$ & $0.170$ & 0.603 & 0.575 & 0.509 & 0.572 \\
energy & 0.732 & $-1.456$ & $-1.471$ & 0.951 & $-1.060$ & $-0.552$ & 0.681 & 0.795 & 0.674 & 0.837 \\
concrete & 0.572 & $-0.331$ & $-0.269$ & 0.841 & $-0.403$ & $-0.235$ & 0.605 & 0.519 & 0.598 & 0.527 \\
wine & 0.423 & $-0.088$ & $-0.158$ & 0.795 & $0.050$ & $0.047$ & 0.548 & 0.577 & 0.583 & 0.494 \\
kin8nm & 0.71 & $-0.326$ & $-0.308$ & 0.884 & $-1.015$ & $-0.887$ & 0.486 & 0.533 & 0.494 & 0.537 \\
power-plant & 0.16 & $-0.114$ & $-0.174$ & 0.526 & $0.147$ & $0.040$ & 0.486 & 0.465 & 0.472 & 0.444 \\
naval & 0.891 & $-0.773$ & $-0.803$ & 0.972 & $-30.2$ & $-12.7$ & 0.913 & 0.741 & 0.958 & 0.726 \\
protein & 0.602 & $-0.131$ & $-0.365$ & 0.866 & $0.230$ & $0.187$ & 0.565 & 0.620 & 0.599 & 0.425 \\
Year-MSD & 0.567 & $-0.358$ & $-0.371$ & 0.799 & $-0.254$ & $-0.334$ & 0.546 & 0.638 & 0.523 & 0.626 \\
superconduct & 0.868 & $-0.624$ & $-0.579$ & 0.981 & $-0.858$ & $-0.659$ & 0.693 & 0.686 & 0.692 & 0.666 \\
sarcos & 0.543 & $-0.758$ & $-0.741$ & 0.876 & $-2.332$ & $-2.077$ & 0.555 & 0.592 & 0.554 & 0.591 \\
diamonds & 0.949 & $-0.849$ & $-0.693$ & 0.997 & $-1.635$ & $-1.408$ & 0.746 & 0.739 & 0.742 & 0.744 \\
video\_transcoding & 0.987 & $-1.020$ & $-0.618$ & 0.998 & $-3.189$ & $-1.632$ & 0.865 & 0.787 & 0.876 & 0.790 \\
medical\_charges & 0.321 & $-0.453$ & $-0.423$ & 0.661 & $-0.988$ & $-0.946$ & 0.508 & 0.523 & 0.505 & 0.544 \\
black\_friday & 0.694 & $-0.001$ & $0.001$ & 0.959 & $0.006$ & $-0.102$ & 0.498 & 0.524 & 0.486 & 0.551 \\
particulate-matter & 0.147 & $-0.457$ & $-0.453$ & 0.337 & $-0.874$ & $-0.854$ & 0.523 & 0.516 & 0.518 & 0.509 \\
nyc-taxi-green & 0.785 & $0.264$ & $0.286$ & 0.959 & $0.571$ & $0.676$ & 0.669 & 0.633 & 0.751 & 0.635 \\
\midrule
bone age & 0.338 & $-0.072$ & $-0.271$ & 0.542 & $-0.621$ & $-1.030$ & -- & -- & -- & -- \\
IMDB-WIKI & 0.006 & $-0.167$ & $-0.197$ & 0.082 & $-0.396$ & $-0.470$ & -- & -- & -- & -- \\
\midrule
median (tabular) & 0.648 & $-0.34$ & $-0.37$ & 0.872 & $-0.62$ & $-0.55$ & 0.584 & 0.606 & 0.591 & 0.581 \\
\bottomrule
\end{tabular}%
}
\end{center}
\end{table}

\section{Cost}
\label{app:cost}
\spp{} adds exactly two things to a IPIV model: the gate heads and the auxiliary triple. For a tabular backbone with hidden width $h$, each gate costs $16h + 65$ parameters ($16h+16$ for the hidden layer, $32$ for batch normalization, $17$ for the output unit) and the auxiliary triple costs $3h + 3$. On Concrete ($h = 50$, $d = 8$) that takes a $603$-parameter IPIV model to $1{,}621$ for the single gate and $2{,}486$ for the dual gate; on superconduct ($h = 100$, $d = 79$) it takes $8{,}303$ to $11{,}936$, a $44\%$ increase. The relative cost falls as the backbone grows and is negligible for the image tasks: on the bone-age VGG-16 stack each gate reads a $1024$-unit representation for $16{,}449$ parameters against a backbone of roughly $15$M, so the two gates and the auxiliary triple together add about $0.2\%$.

Training cost is the more meaningful number, and we measure it directly rather than estimate it. Across the nine UCI datasets, training one \spp{} model takes a median $\mathbf{1.63\times}$ the wall-clock of training one IPIV model on the same data and hardware (range $1.00\times$ on Yacht to $2.03\times$ on Protein, the largest). The whole-pipeline cost is higher than this ratio suggests, because \spp{} trains one model per acceptance rate: a 10-point curve costs roughly $16\times$ a single IPIV fit.

Inference is where the architecture pays off. Both gates are scalar heads on the shared trunk, so an accept/reject decision, an interval and a point estimate all come from \emph{one} forward pass --- the same cost as plain IPIV, and unlike an ensemble or MC-dropout selector, which need $K$ or $T$ passes for their uncertainty signal. The deep-ensemble selector of \S\ref{sec:setup} is the concrete case: it costs five IPIV fits per split to train and five forward passes per query at inference, against one \spp{} fit per acceptance rate and one pass per query; the comparison in Table~\ref{tab:auroc} (right columns) is made at that fivefold inference cost.

\section{Additional Metrics Discussed in the Literature}
\label{app:cwc}
The coverage-width criterion of the LUBE/QD line combines coverage and width into one number, $\mathrm{CWC} = \mathrm{NMPIW}\,\big(1 + \gamma(\picp)\,e^{-\eta(\picp - \mu)}\big)$, where $\mathrm{NMPIW}$ is the mean width divided by the target range, $\mu$ is the nominal level, and $\gamma$ is $1$ only when coverage falls below nominal. We computed it for completeness but do not headline it, for two reasons. It is not a proper scoring rule, so it cannot be used to rank forecasters without the ranking depending on $\eta$; and its exponential penalty makes it extremely sensitive near the nominal level, which shows plainly in our numbers --- on wine, a drop in selective coverage from $0.838$ to $0.791$ moves CWC from $86$ to $605$, a sevenfold change from a five-point coverage difference. The interval score of \eqref{eq:winkler} measures the same trade-off with a linear miss penalty and a decision-theoretic justification, which is why it is our headline.

\end{document}